%% file: workshop_main.tex
\documentclass{article}

\usepackage[final]{workshop_neurips/mlsafety_neurips_2022}

\usepackage{mdframed}
\usepackage[utf8]{inputenc} 
\usepackage[T1]{fontenc}    
\usepackage{amsmath}
\usepackage[colorlinks=true,linkcolor=blue,citecolor=blue, pagebackref=true]{hyperref} 
\usepackage[capitalise, nameinlink, noabbrev]{cleveref}       
\renewcommand*\backref[1]{\ifx#1\relax \else (Cited on #1) \fi}
\usepackage{url}            
\usepackage{booktabs}       
\usepackage{amsfonts}       
\usepackage{nicefrac}       
\usepackage{microtype}      
\usepackage[dvipsnames]{xcolor}         

\usepackage{graphicx}
\usepackage{subfigure}

\usepackage{amssymb}
\usepackage{amsthm}
\usepackage{enumitem}
\usepackage{thm-restate}
\usepackage[ruled,vlined]{algorithm2e}
\usepackage[normalem]{ulem}
\usepackage{comment}
\usepackage{wrapfig}
\usepackage{array}
\usepackage{float}

\usepackage{todonotes}

\DeclareUnicodeCharacter{2212}{−}
\def\dgm{\mathop{\sf PD}}
\def\SW{\mathop{\sf SW}}

\theoremstyle{plain}
\newtheorem{theorem}{Theorem}

\newtheorem{lemma}[theorem]{Lemma}

\theoremstyle{definition}

\theoremstyle{remark}

\crefname{algocf}{Algorithm}{Algorithms}
\Crefname{algocf}{Algorithm}{Algorithms}

\title{An Adversarial Robustness Perspective on the
Topology of Neural Networks}

\author{%
  Morgane Goibert  \\
  Criteo AI Lab \\
  Paris, France \\
  \texttt{m.goibert@criteo.com} \\
   \And
   Thomas Ricatte \\
   Amazon \thanks{Work done at Criteo AI Lab} \\
   Luxembourg \\
   \texttt{tricatte@amazon.com} \\
   \And
   Elvis Dohmatob \\
   Meta AI Research \footnotemark[1] \\
   Paris, France \\
   \texttt{dohmatob@fb.com} \\
}

\begin{document}

\maketitle
\addtocontents{toc}{\protect\setcounter{tocdepth}{0}}

\input{workshop_neurips/abstract}
\input{workshop_neurips/introduction.tex}
\input{workshop_neurips/hypothesis.tex}
\input{workshop_neurips/methods.tex}
\input{workshop_neurips/results.tex}
\input{workshop_neurips/conclusion.tex}

\clearpage
\nocite{*}
\bibliographystyle{plain}
\bibliography{workshop_main}


\clearpage
\addtocontents{toc}{\protect\setcounter{tocdepth}{2}}
\input{workshop_neurips/supplementary.tex}

\end{document}

%% file: workshop_neurips/abstract.tex
\begin{abstract}
In this paper, we investigate the impact of neural networks (NNs) topology on adversarial robustness. Specifically, we study the graph produced when an input traverses all the layers of a NN, and show that such graphs are different for clean and adversarial inputs. We find that graphs from clean inputs are more centralized around highway edges, whereas those from adversaries are more diffuse, leveraging under-optimized edges. Through experiments on a variety of datasets and architectures, we show that these under-optimized edges are a source of adversarial vulnerability and that they can be used to detect adversarial inputs.
\end{abstract}

%% file: workshop_neurips/introduction.tex
\section{Introduction}

As neural networks (NNs) can be fooled by adversarial examples~\citep{szegedy2013intriguing, goodfellow2014explaining}, understanding them remains an important issue. Several hypotheses have been proposed to explain this still obscure phenomenon~\cite{ilyas2019adversarial, nakkiran2019a,xu2019interpreting,rice2020overfitting,backdoor,stutz2019disentangling}. In this paper, we build on identified characteristics of adversarial examples to make the following hypothesis: adversarial inputs take different paths than clean inputs when they traverse a neural network (NN), namely under-optimized edges. To study this hypothesis, we will use notions from the adversarial robustness field, but also from topological data analysis.

\paragraph{Adversarial Examples.}
An \textit{adversarial example} is a perturbed version of a clean input $x$, i.e $x^{adv} = x + \delta$, where $\delta$ is the perturbation controlled in size ($L_2$ or $L_\infty$ norm, say) by a strength parameter $\varepsilon$. An \emph{attacker} is any mechanism that constructs such example to cause a given classifier $h$ to misclassify the example: $h(x^{adv}) \neq h(x)$, in which case the attack is called successful. 

\paragraph{Topological Data Analysis.}
Topological Data Analysis (TDA)~\cite{edelsbrunner2000topological, zomorodian2005computing} is a field which uses tools from ideas from topology to analyze high-dimensional data like graphs~\cite{li2014persistence, carriere2015stable, turner2014persistent}. TDA is well-suited for studying the structural properties of data while reducing the dimension of the analysis, which fits our case since our data are high-dimensional and associated with activation graphs from NNs. Our paper critically relies on \textit{persistence diagrams}, which summarize the topological structure of weighted graphs with a set of points in $\mathbb{R}^2$.

\paragraph{Contributions.} The main aim of this paper is to demonstrate that the analysis of the topological structure of NNs is highly relevant to better understand, detect, and defend against the adversarial phenomenon. We pave the way for this new line of work in this paper, which is organized as follows:
\begin{itemize}[noitemsep, leftmargin=8pt, topsep=0pt]
    \item In \cref{sec:hypothesis}, we propose a new hypothesis on how the topological structure of NNs and under-optimized parameters are related to the adversarial phenomenon.
    \item In \cref{sec:methods}, we propose main method to extract structural topological features based on \textit{persistence diagrams} and under-optimized edges.
    \item In \cref{sec:preliminary_exp} We conduct experiments to validate our hypothesis using our newly-defined features.
\end{itemize}

%% file: workshop_neurips/hypothesis.tex
\section{Our hypothesis}
\label{sec:hypothesis}

\begin{wrapfigure}[15]{r}{0.36\textwidth}
    \centering
    \vspace{-0.2cm}
    \includegraphics[width=0.35\textwidth]{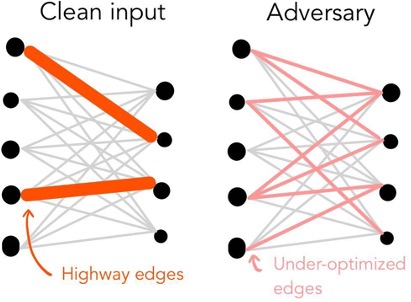}
    \caption{Blueprint of structural differences between graphs from clean vs adversarial inputs.}
    \label{fig:schema_hypothesis}
\end{wrapfigure}

Based on the observation that most NNs are over-parametrized (i.e parameter count exceeds training dataset size) and that pruning away most parameters after training induces smaller models without degrading accuracy~\cite{frankle2018lottery}, we hypothesize that only a small set of parameters are critically used for inference of clean inputs, while the rest of the parameters do not carry meaningful information. Considering a NN as a graph, and parameters as edges of that graph, this means that information from clean inputs flows through highway edges, while information from adversarial inputs is more diffuse, and uses so-called under-optimized edges (i.e. useless edges not well optimized during training). This results in \textit{structural differences} in graphs induced by clean and adversarial inputs, as simply illustrated by \cref{fig:schema_hypothesis}. Using the notion of \textit{induced graph}, which is a weighted graph representing the information flow from an input in a NN/graph, and defined later, we can sum up our hypothesis:

\begin{mdframed}
\textbf{Our Hypothesis.} \emph{Clean and adversarial inputs induce differences in the topological structure in their respective induced graphs, because under-optimized edges are used by adversaries, but not by clean inputs. Such edges are thus a source of adversarial vulnerability.}
\end{mdframed}

%% file: workshop_neurips/methods.tex
\section{Extracting Structural Topological Features: Methods}
\label{sec:methods}

\paragraph{Induced Graph.}
\label{sec:induced_graph_threshold}

Let $\mathcal X=\mathbb R^{n_0}$ be the feature space, where $n_0$ is the input dimension. For any input $x \in \mathcal{X}$, the induced graph (aka activation graph) is a graph on the neurons of the network, whose edges depend both on the parameters of the network and the inner activations induced by the forward pass of $x$.
Formally, a NN is a function $h:\mathcal X \rightarrow [\![1,K]\!]$ of the form $h(x) = \arg\max_{k=1}^K g(x)_k$ where the feature map $g: \mathcal X \rightarrow \mathbb R^K$ is defined by $g(x) = \sigma_L(W_L \, \sigma_{L-1}(W_{L-1} \ldots (\sigma_1(W_1 \, x)))$, where $W_l \in \mathbb R^{n_l \times n_{l-1}}$ is the parameter matrix between layer $l-1$ and layer $l$; the component-wise mappings $\sigma_l: \mathbb R \rightarrow \mathbb R$ are the activation functions of the NN (e.g. ReLU). With a slight abuse of notation, we denote by $g_l(x) \in \mathbb R^{n_l}$ the output value of layer $l$.

Combining information from the NN $g$ and an input $x$, we construct the so-called \textit{induced graph}.
\begin{equation*}\begin{split}
&G(x,g) = (V,E),\text{ with }V=\{1,2,\ldots,n_0+\ldots +n_L\} \\
&\text{ and }E = \{(u^l,v^{l+1},w^{l}_{u,v})\} \subseteq V^2 \times \mathbb R.                  
\end{split}\end{equation*}

Here, the edge weights are given by $w^l_{u,v} = | [g_l(x)]_u \times (W_l)_{v,u} |$, the value of the parameter weight of the NN between neurons $u$ and $v$ multiplied by the activation of neuron $u$: this definition of $w^l_{u,v}$ is meant to mimic how NNs operate to transfer information from a layer to the next. 
It applies to feedforward NNs, and can also be generalized for other structures like ResNet. Moreover, the $w_{u,v}$'s can also be obtained for convolutional layers or others (see details in \cref{sec:suppl_persistent_dgm_comput}).

\paragraph{Selecting under-optimized edges.} As classical NNs have a huge number of parameters (even for small ones as LeNet), it is necessary to reduce dimensionality and select a sub-graph of the induced graph. Moreover, as we expect adversaries to leverage \textit{under-optimized} edges, we select only these edges for our analysis. As defined and studied in~\cite{frankle2018lottery,zhou2019deconstructing}, an edge $(u,v)$ is under-optimized if the \textit{Magnitude Increase} (MI) quantity $ |(W_l)_{u,v}| - |(W_l^{init})_{u,v}|$ is small, $(W_l^{init})_{u,v}$ being the parameter's initialization value.
An edge $(u,v)$ of layer $l$ is kept in the thresholded induced graph iff:
\begin{equation}
    |(W_l)_{u,v}| - |(W_l^{init})_{u,v}| < \text{quantile}(q)\;\;,
\label{eq:threshold_param}
\end{equation}
where $q$ is the target fraction of edges to keep. We denote the \textit{thresholded induced graph} as $G^q(x,g)$. Note that no assumption is made over the initialization of the NN and that the selection criterion of under-optimized edges does not depend on the input $x$, but only on the NN $g$.

\paragraph{Persistent Homology.}
\label{subsec:persistent_homology}
We can analyze our under-optimized induced graph structure using TDA tools, namely persistent homology. We only provide a simple overview and some intuitions about the concepts we use, but the interested reader can find more details in \cite{chazal2017introduction} and \cref{sec:suppl_persistent_dgm_comput}. We are interested in an object called a persistent diagram (of dimension $0$ in our case): it is a set of points representing the topological structure of a graph through different spatial scales. The graph is decomposed into a sequence of sub-graphs, starting with a completely disconnected graph, and ending with the whole graph. In between, edges from the graph are added progressively according to their weights (highest weights first). The evolution of connected components in the sequence of sub-graphs is tracked with two indicators: the birth date of the connected component (when a first edge appears), and its death date (when the connected component is linked to another, older one, through an edge appearing at this death date, or $+\infty$ when the connected component never dies). This collection of points (birth dates and death dates) is the persistence diagram, abbreviated PD. Intuitively, we can derive some simple observations. A highly connected graph, with weights very close to each other, will have very few points in the PD, and only one infinitely-lived point. On the contrary, a disconnected graph, with very different weights for the edges, will have many points and infinitely-lived points in its PD. 

Thus, PDs are very well suited to studying the topological structure of a graph. We expect PDs from clean inputs to have fewer points / fewer infinitely-lived points than those from adversaries.

%% file: workshop_neurips/results.tex
\section{Clean and Adversarial Examples induce Different Persistence Diagrams}
\label{sec:preliminary_exp}

\begin{figure}[h!]
\centering
    \subfigure[Distribution of all PD points.]{\label{fig:viz_distrib_pts_dgms}\includegraphics[width=0.47\textwidth]{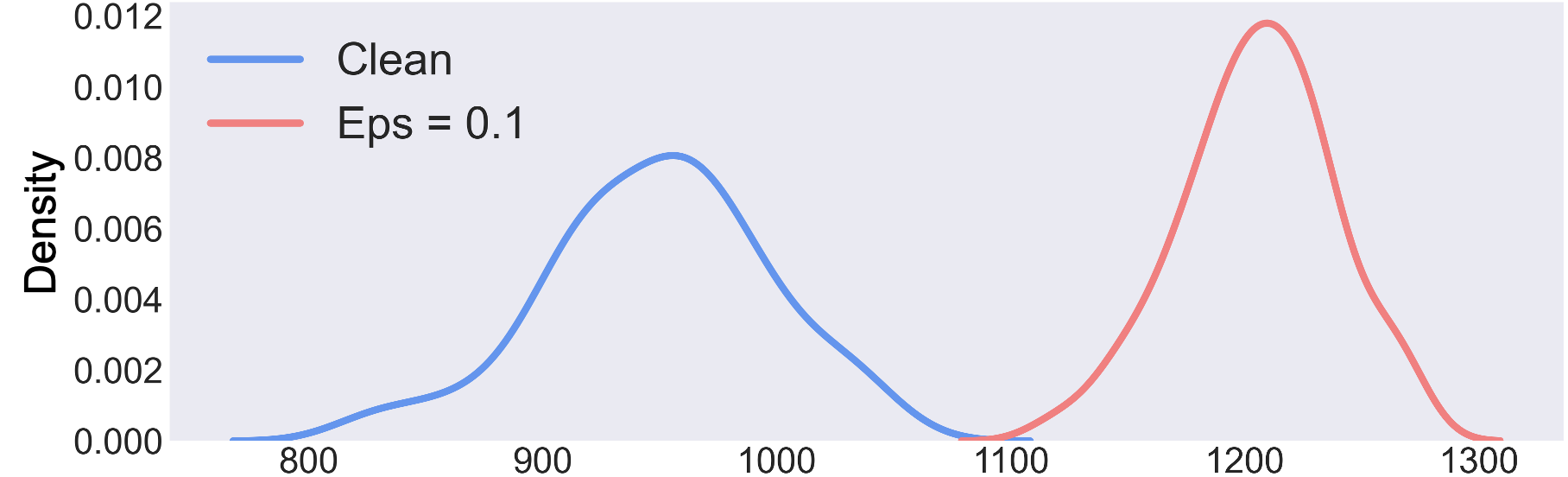}}
    \hspace{0.5cm}
    \subfigure[Distribution of infinitely-lived PD points.]{\label{fig:viz_distrib_pts_dgms_inf}\includegraphics[width=0.47\textwidth]{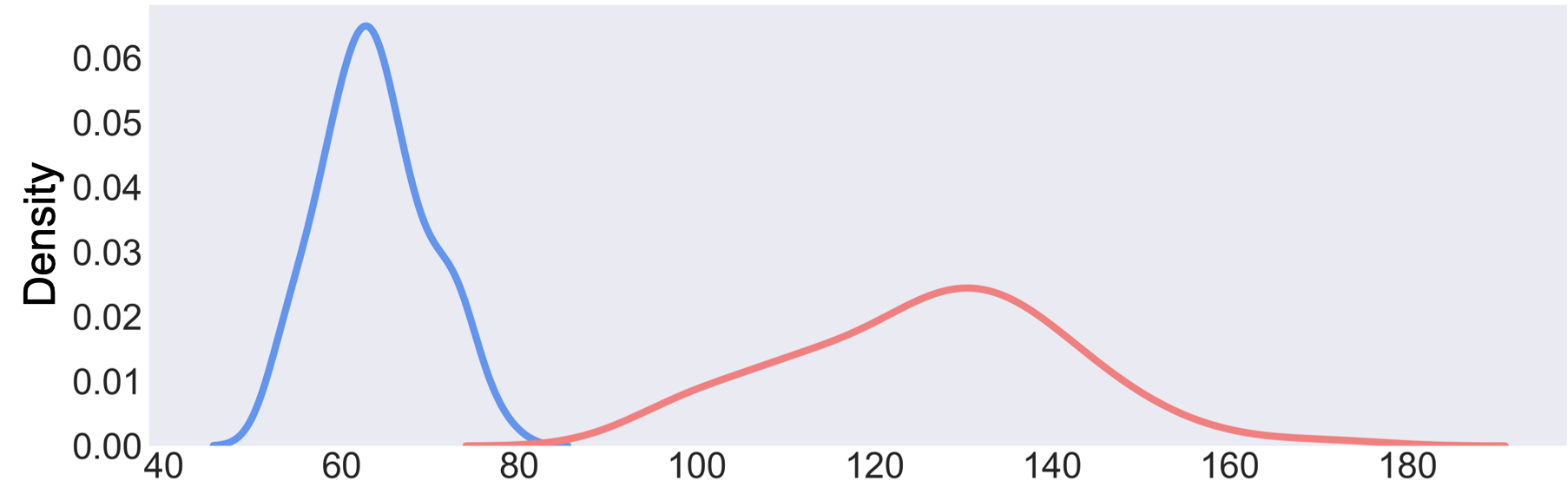}}
    \caption{PD points computed on MNIST / LeNet}
    \label{fig:viz_distrib_pts_dgms_all}
\end{figure}

\paragraph{Observing Quantitative Differences.}
When the induced graphs are sufficiently small, differences in PDs can be easily observable based on the number of points in the PDs. \cref{fig:viz_distrib_pts_dgms_all} shows this is the case for a classical MNIST / LeNet, where adversaries were computed using PGD~\cite{kurakin2016adversarial} with $\varepsilon=0.1$.

\paragraph{Detecting Adversarial Examples.} While differences in PDs are easily observable on simple setups, it is necessary to extend our analysis to more complex, SOTA setups. Even though not as easily observable in these cases, we derived a detection framework based on PDs, which can be used for any dataset/architecture, whose success shows that adversarial PDs (and thus adversarial inputs) are indeed different from clean ones, for a variety of SOTA attacks (PGD~\cite{kurakin2016adversarial} and CW~\cite{carlini2017towards} for the white-box setting, Boundary ~\cite{brendel2017decision} for the black-box one) and datasets (MNIST, Fashion MNIST, SVHN, CIFAR10), using LeNets and ResNets architectures. Our code is available at: \url{https://github.com/detecting-by-dissecting/detecting-by-dissecting}.

We defined a feature extraction map for our so-called PD method: $\Phi_{\dgm}(x,g) := \dgm(G^q(x,g))$. To compute distances between different PDs, we used the Sliced Wasserstein Kernel, defined in~\cite{carriere2017sliced} by: $K_{\dgm}(x,x') = \exp\left(-\frac{1}{2\sigma^2}\SW(\Phi_{\dgm}(x,g), \Phi_{\dgm}(x',g))\right)$, where $\SW(\cdot,\cdot)$ is the Sliced-Wasserstein distance between persistence diagrams.

Based on this PD-based feature extraction method and a kernel, we can build a detector using a simple SVM. We compare our method, called PD (for simplicity), to SOTA detection baselines: \emph{Mahalanobis}~\citep{lee2018simple} and \emph{Local Intrinsic Dimension (LID)}~\citep{ma2018characterizing}. For the sake of comparison, we also compare our PD method to a very simple one called \emph{Raw Graph (RG)}, whose features are just a vector whose elements are the weights of the thresholded induced graphs $G^q(x,g)$.

\begin{figure*}[ht!]
    \centering
    \subfigure{\label{fig:legend}\includegraphics[width=0.7\linewidth]{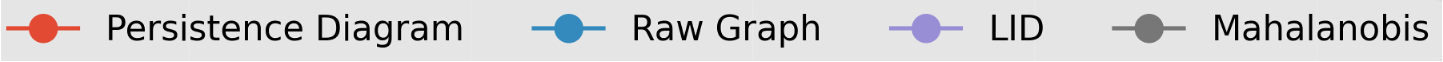}}
    \setcounter{subfigure}{0}
    
    \centering
    \subfigure[LeNeT/MNIST]{\label{fig:res_mnist_lenet}\includegraphics[width=0.245\textwidth]{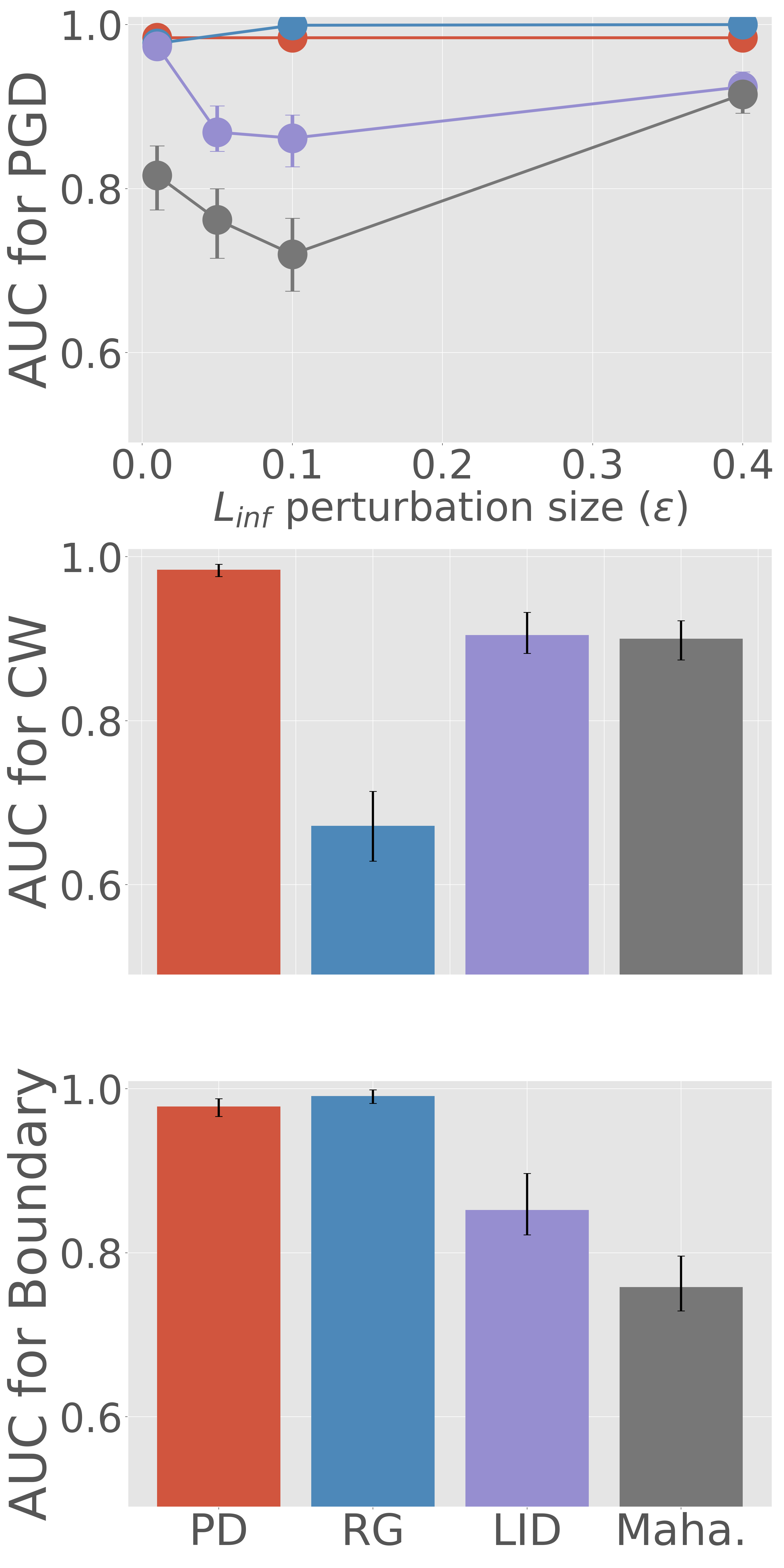}}
    \subfigure[LeNet/Fashion MNIST]{\label{fig:res_fashionmnist_lenet}\includegraphics[width=0.245\textwidth]{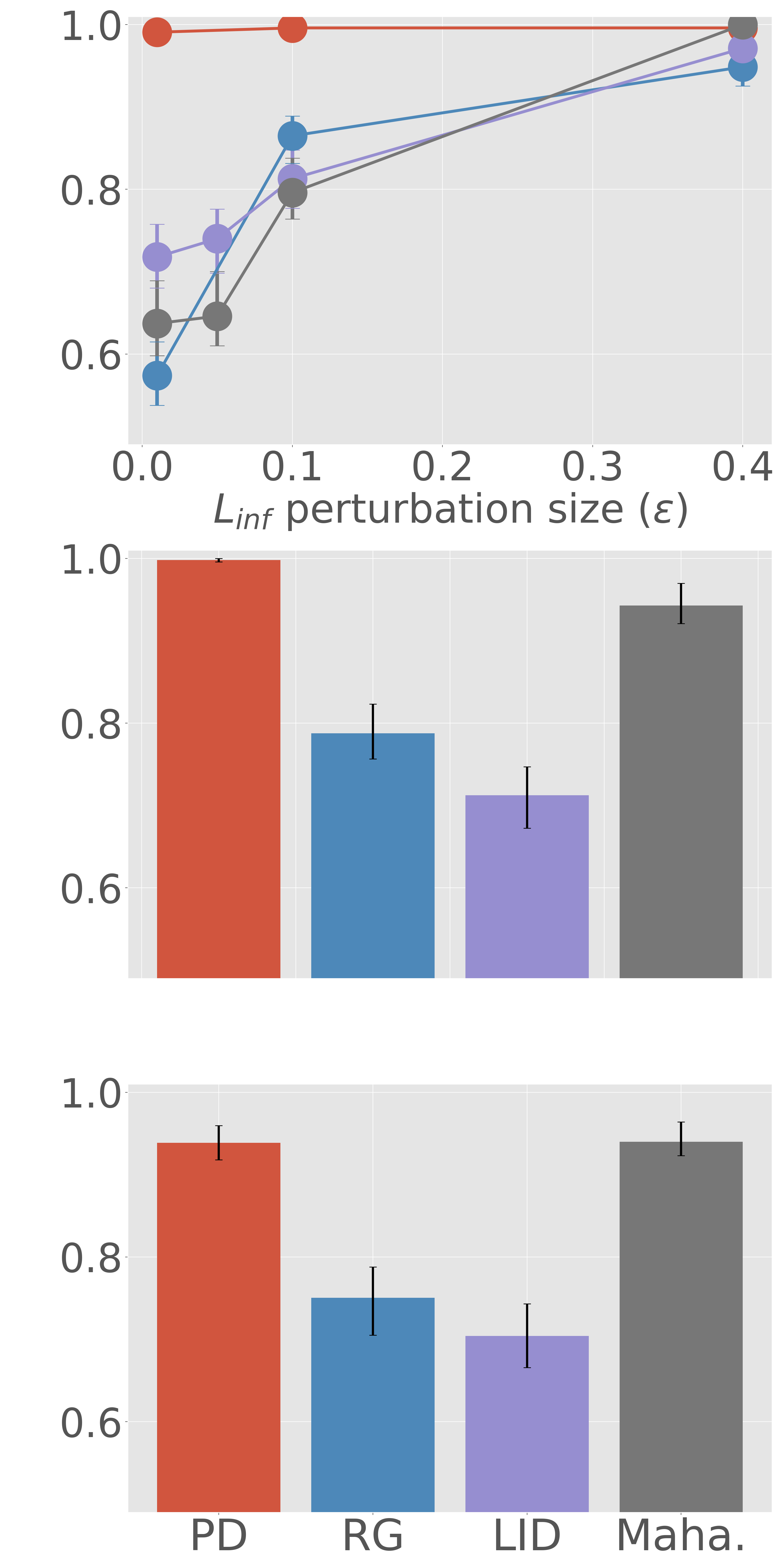}}
    \subfigure[ResNet/SVHN]{\label{fig:res_svhn_resnet}\includegraphics[width=0.245\textwidth]{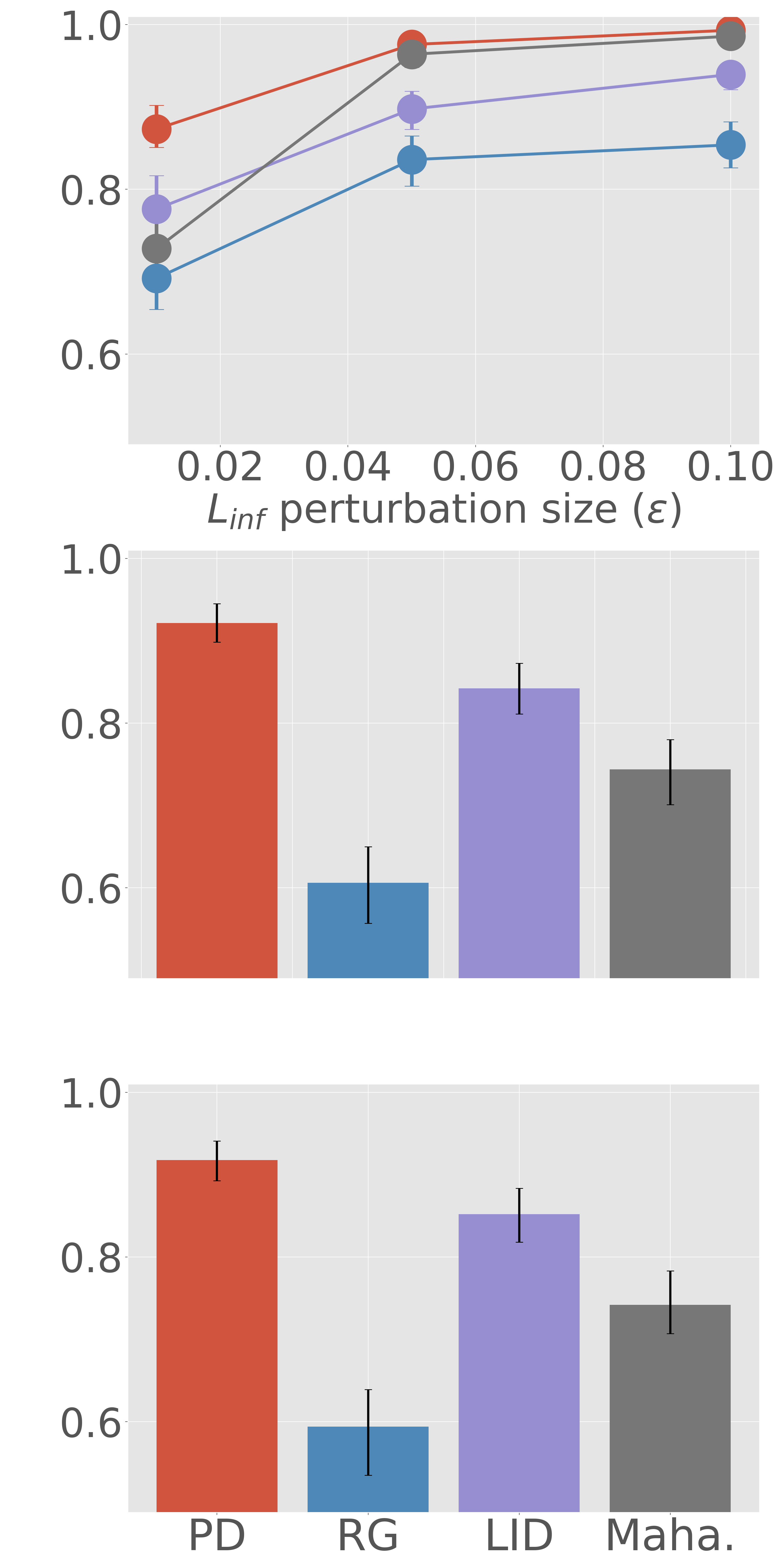}}
    \subfigure[ResNet/CIFAR10]{\label{fig:res_cifar_resnet}\includegraphics[width=0.245\textwidth]{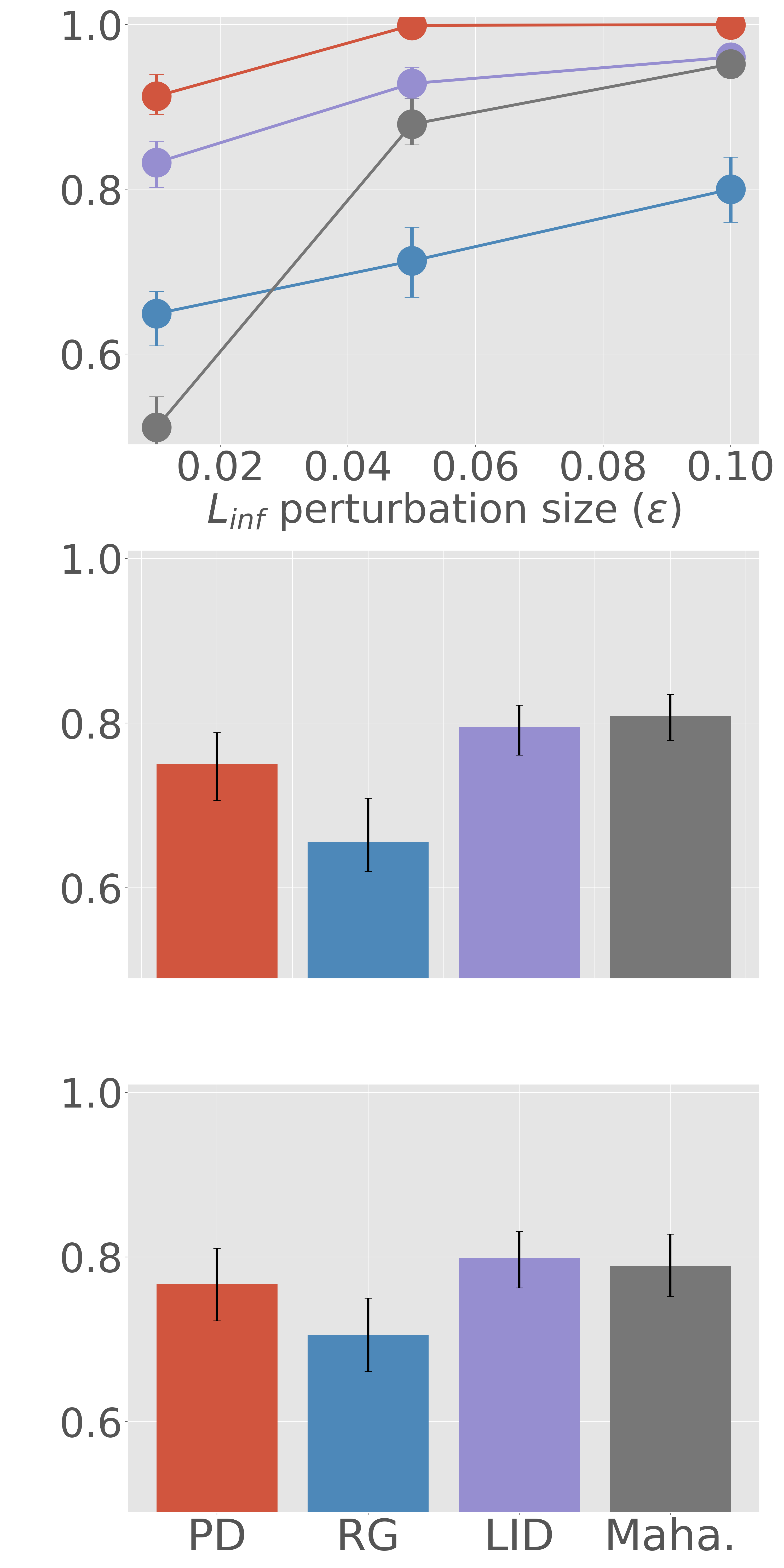}}
\caption{Showing detection AUC for different detection methods (legend) against different kinds of adversarial attacks (rows) and model architectures and datasets (columns). We see that our proposed method based on PD outperforms the SOTA methods, except for one tie.}
\label{fig:principal_exp_results}
\end{figure*}

\cref{fig:principal_exp_results} presents the AUC detection results for the different methods, against our three attacks and four setups. PD has better AUC results than SOTA methods on the four datasets/architectures and on all attacks, except on CIFAR10 ResNets, where the results are similar. RG remains competitive with the two baselines on the (small) LeNet architectures. The main takeaways of these experiments are:

\begin{itemize}[noitemsep, leftmargin=7pt, topsep=0pt]
    \item RG's performances indicate that useful information can indeed be found in the thresholded induced graph, thus in the under-optimized edges. However, such a simple method is only efficient on simple models or attacks.
    
    \item PD's performances are overall significantly better than those of previous SOTA detectors, LID and Mahalanobis. This means not only we have succeeded at constructing a very effective detector, but also that structural topological information extracted from induced graphs contains discriminative information about adversarial examples, regardless of the task complexity. Overall, the success of PD validates our main hypothesis.
\end{itemize}

The results on the Boundary black-box attack show that our methods (and also the baselines LID and Mahalanobis) do not rely on gradient masking and can generalize well. More experiments and illustrations on PDs and under-optimized edges are provided in \cref{sec:additional_results,sec:how_pd_works,sec:adv_training_exp}.

%% file: workshop_neurips/conclusion.tex
\section{Conclusion}
\label{sec:conclusion}
\paragraph{Summary.} We studied the topological structure of NNs through the lens of adversarial robustness. We stated that clean and adversarial inputs follow different paths when they traverse a NN, resulting in different topological structures for their induced graphs. Namely, contrary to clean inputs, adversarial ones leverage under-optimized edges, whose existence stems from the over-parametrization of NNs. We verified this hypothesis through a variety of experiments.

\paragraph{Takeaways.} Our paper is, to the best of our knowledge, one of the first to link adversarial robustness and the topological structure of NNs. We validate the need for more in-depth analysis and understanding of the topological structure of NNs, of how the information from an input $x$ flows through a NN, and of the impact of over-parametrization on deep learning. In the context of adversarial robustness, these lines of research are still not explored, but can greatly improve our understanding of the phenomenon.

\paragraph{Future works.} Refinements and additional experiments on our PD method are left for future work. Moreover, a better understanding of under-optimized edges (e.g. their trajectories during training, etc.), and the study of the link between pruning (i.e. removing under-optimized edges) and adversarial robustness are also left for future work.

%% file: workshop_neurips/supplementary.tex
\appendix
\setcounter{section}{0}

\begin{center}
    \noindent\rule{\textwidth}{4pt} \vspace{-0.2cm}
    
    \Large \textbf{Supplementary Material} \\ ~\\[-0.5cm]
    \large \textbf{An Adversarial Robustness Perspective on the Topology of Neural Networks} 
    
    \noindent\rule{\textwidth}{1.2pt}
\end{center}

\tableofcontents
\clearpage

\section{Characteristics of Adversaries: Motivation Details on Under-Optimized Edges}

\label{sec:understanding_adv}
Adversarial perturbations are small and yet result in sufficient variation of the output to change the predicted class. What happens inside a NN to obtain this variation? We recall here three characteristics of adversaries and link them together to suggest an answer to this question and motivate the use of graphs and topological tool to study adversaries.

\paragraph{Strategies used by adversaries.}
\cite{xu2019interpreting} shows that adversarial perturbations can be categorized into \textit{suppressing} ones, meaning perturbations that focus on reducing the true label score, or \textit{promoting} ones, meaning perturbations that focus on increasing the target label score. Adversaries can (and usually do) output a mixed behavior. Interestingly, the suppressing/promoting nature of an adversary comes from the set of input features (e.g. pixels for images) it perturbs: modification in one input neuron cascades through the whole NN and results in a suppressing/promoting relative behavior.

\paragraph{What features are used by adversaries?} Using \cite{ilyas2019adversarial, nakkiran2019a} terminology, the features of the data distribution can be divided into 1) useful and robust, 2) useful and non-robust, 3) non-useful ones. Both of these works show the existence of two types of adversaries (see also \cite{stutz2019disentangling}), even though one can expect that most adversaries lie on a scale between these two extremes:
\begin{itemize}[noitemsep, leftmargin=8pt, topsep=0pt]
    \item Adversaries leveraging useful and non-robust directions: e.g. when an image from the class "dog" is perturbed to be classified as a "cat", the perturbation has something to do with the class "cat". Then, the adversary is on-distribution (the direction of the perturbation is parallel to the data manifold, thus the adversary does not leave the data manifold).
    \item Adversaries leveraging non-useful directions: e.g. the image from class "dog" is perturbed with a perturbation that has nothing to do with class "cat". Then, the adversary is off-distribution because the perturbation can occur in any arbitrary direction (the direction of the perturbation is perpendicular to the data manifold, thus the adversary leaves the manifold).
\end{itemize}

\paragraph{Over-parametrization.} The link between over-parametrization and robustness is still not completely understood, however, some works (e.g. \cite{rice2020overfitting,backdoor,widerwu2021}) have shown that NNs vulnerability may increase when they are over-parametrized. It occurs when a NN has too many parameters: after training with e.g. SGD, parameters in excess still have non-zero values, and thus are used for prediction. \begin{wrapfigure}[12]{l}{0.48\textwidth}
    \centering
    \vspace{-0.2cm}
    \includegraphics[width=0.47\textwidth]{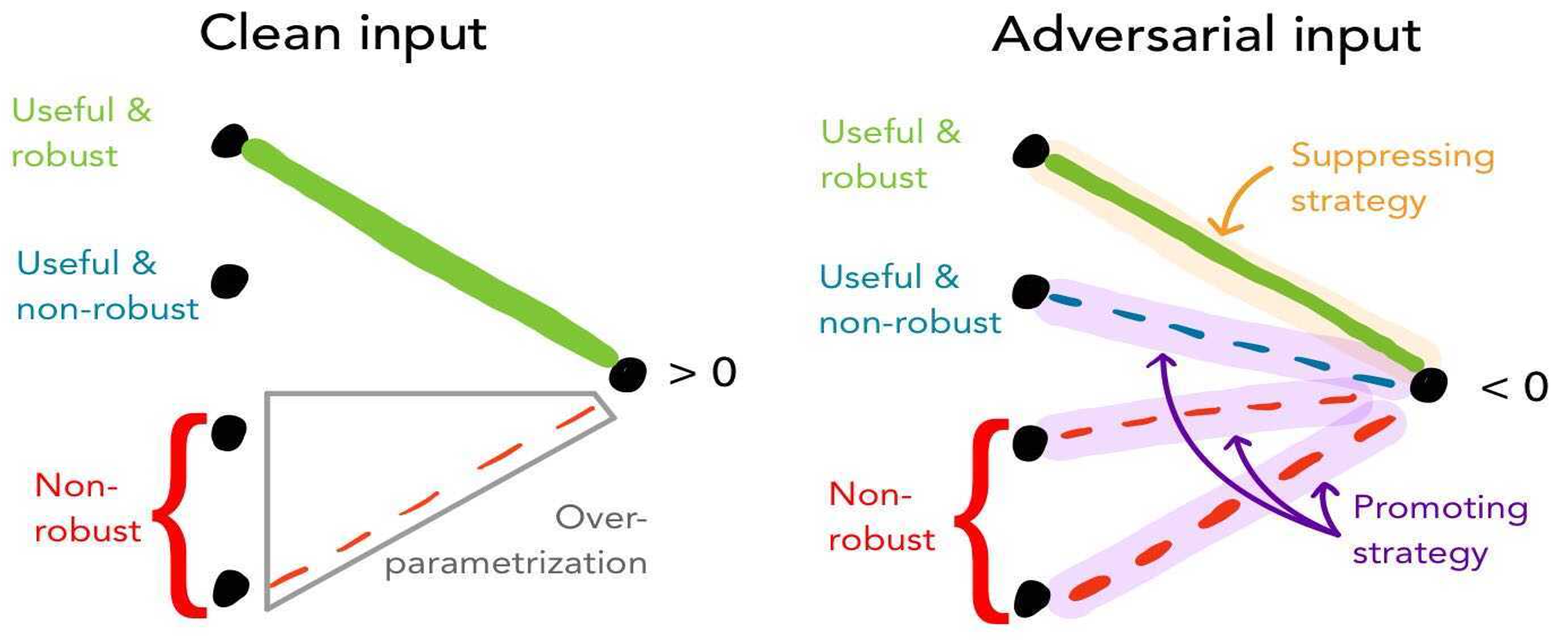}
    \caption{Adv. inputs characteristics. Full (dashed) lines denote positive (negative) weights and thickness denotes absolute value.}
    \label{fig:characteristisc_adv_ex}
\end{wrapfigure}
It enables highly curved decision boundaries~\cite{liu2020some} and can lead to overfitting the training data. Thus, over-parametrization can translate into having a NN with many under-optimized and non-useful parameters for the classification task at hand.
These non-useful parameters can be leveraged to build adversarial attacks (e.g. via \textit{promoting} behaviors). Such behavior is the most expected one for standard NNs, because they usually are over-parametrized, and most attacks (e.g. PGD) use non-useful directions to perturb clean inputs~\cite{stutz2019disentangling}.
In the alternative case where under-optimized and non-useful parameters are removed (by e.g. pruning), adversarial perturbations can still leverage useful but non-robust parameters to create on-distribution adversarial examples.

\cref{fig:characteristisc_adv_ex} illustrates these characteristics, leading the NN to classify the clean input (resp. adversarial input) as a positive (negative).

\section{Induced Graphs and Persistence Diagram: Details}
\label{sec:suppl_persistent_dgm_comput}

\subsection{Induced Graphs}

\begin{figure*}
    \centering
    \subfigure[Trained NN]{\label{fig:trained_nn_main}\includegraphics[height=0.25\textwidth]{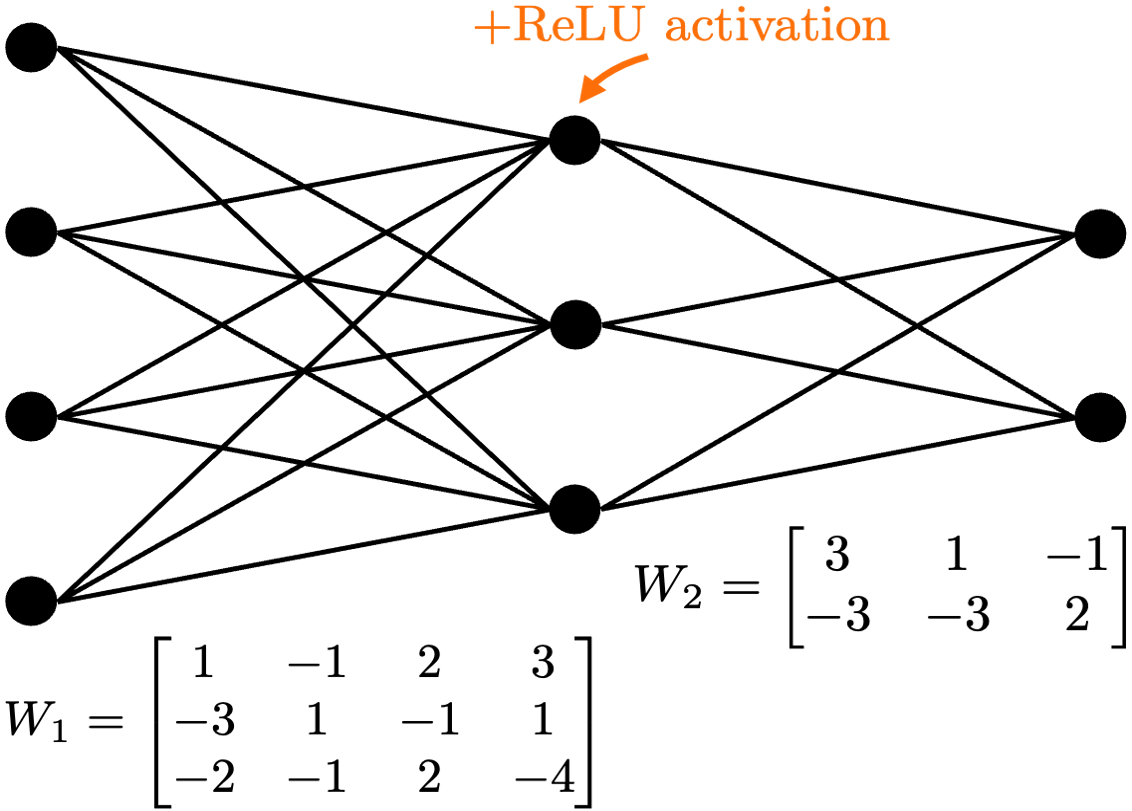}}
    \hspace{0.7cm}
    \centering
    \subfigure[Induced graph]{\label{fig:induced_graph_main}\includegraphics[height=0.25\textwidth]{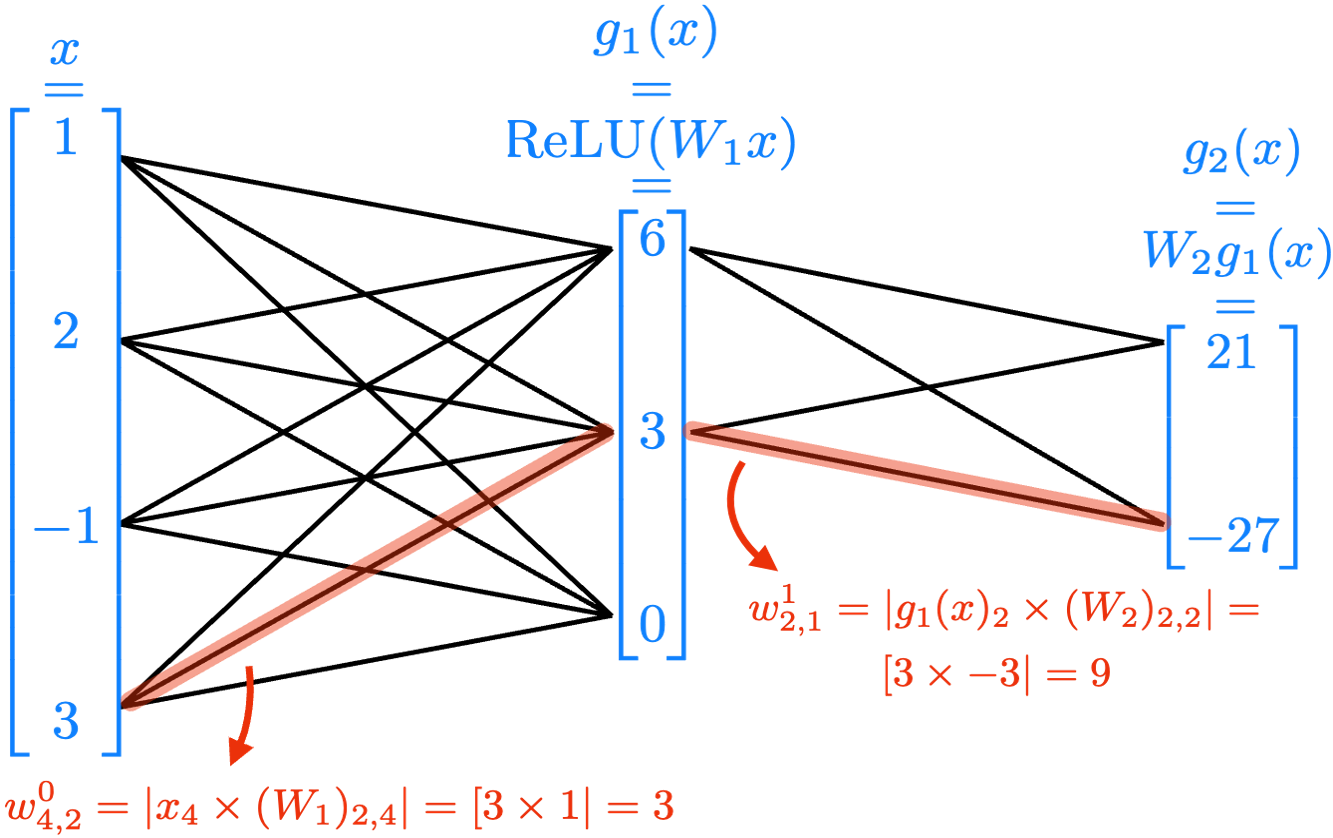}}
\caption{A trained NN (a) and its corresponding induced graph for an input $x$ (b). We highlighted the \textit{activation values} at each layer (blue), i.e. the values of the neurons. We also provided the weights for two edges (red), which denotes the information flow from input $x$ carried by the edge.}
\label{fig:induced_graph_illustration}
\end{figure*}

\cref{fig:induced_graph_illustration} provides an illustration of the way an induced graph is computed. \cref{fig:trained_nn_main} shows a trained NN, with the weights for each layer written in the matrices. For an input $x = (1,2,-1,3)$, \cref{fig:induced_graph_main} shows the corresponding induced graph. Another illustration is provided in \cref{fig:the_dgm_pipeline}

\subsection{Persistence Diagrams: More Intuition}

\paragraph{Simplicial complexes.} A simplicial complex is a topological object generalizing the notion of triangulation, composed of vertices and edges. Up to some constraints, it is a set of simplexes (a $n$-simplex is a triangle in dimension $n$).
We can smoothly compute their homology groups, whose elements, homology classes, represent different structural "holes" and are our relevant topological information. A graph, like our induced graphs, is of course composed of vertices and edges and thus can be seen as a simplicial complex.

\paragraph{Persistence diagrams.} In order to study the topological features at different scales, we decompose the simplicial complex as a \emph{filtration} of sub-complexes. The set of homology groups of each element of the filtration is called a \emph{persistent homology}. A persistence diagram (PD) is the representation of the birth and death dates of the homology classes through the filtration. Then, a point with a long lifetime (far from the diagonal) represents a feature for the simplicial complex under study; on the contrary, a point with a short lifetime (close to the diagonal) represents noise.

\begin{figure*}
    \centering
    \subfigure[A regular graph and its PD]{\label{fig:graph1_pd}\includegraphics[width=0.46\textwidth]{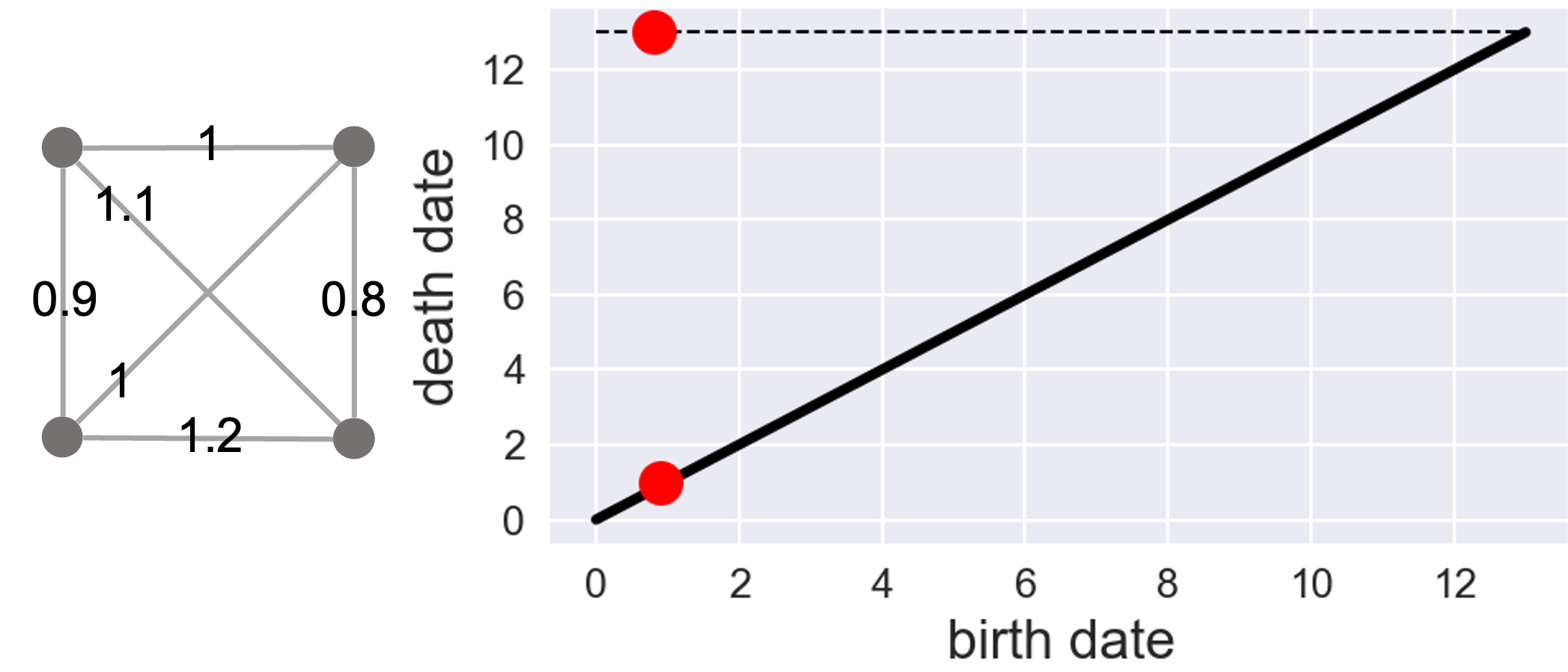}}
    \hspace{0.5cm}
    \centering
    \subfigure[A structured graph and its PD ]{\label{fig:graph2_pd}\includegraphics[width=0.46\textwidth]{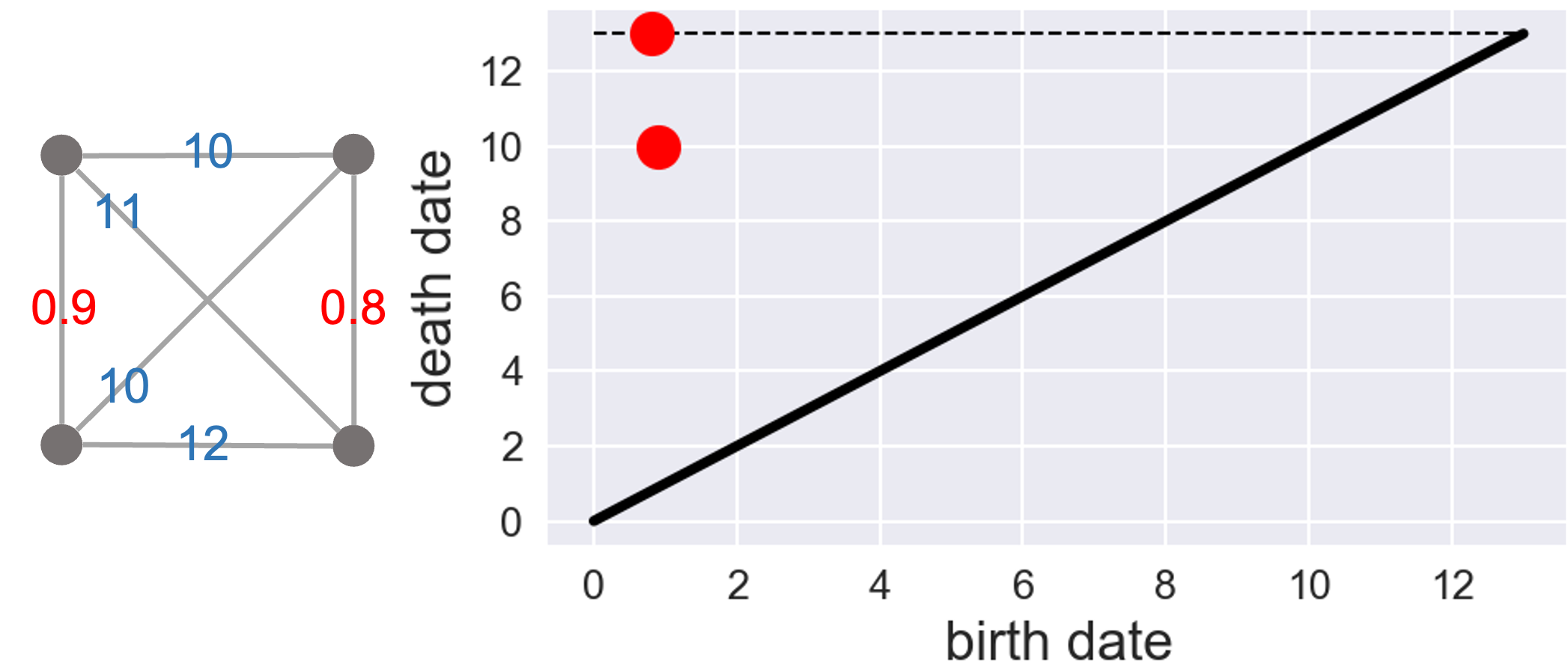}}
\caption{Two graphs with different topological structures and their corresponding PDs (dashed lines correspond to infinity). In (a), the weights are similar: the only important subgraph is the whole graph, thus one point is far from the diagonal. In (b), there are two edges with much smaller values than the others (red): they form two important subgraphs, thus two points far from the diagonal.}
\label{fig:pd_illustration}
\end{figure*}

\paragraph{Intuitions and illustrative example.} As our graphs are feedforward and do not represent 3-d objects, we focus our analysis on the $0^{\text{th}}$-dimensional persistence diagrams. The sub-complex for parameter $t$ thus is the sub-graph composed of edges with weights smaller than $t$ (and corresponding neurons). The filtration is the collection of sub-complexes from $t=0$ (empty graph) to $t=+\infty$ (whole graph). Intuitively, the persistence diagram then represents how the connected components of the sub-complexes evolve through different spatial scales given by the weights of the graph. Highly connected subsets of edges (with small edge weights) will form a connected component during many sub-complexes: it will create a point in the persistence diagram with a long lifetime, far from the diagonal, representing an important structural feature for the whole graph. An illustration is given in \cref{fig:pd_illustration}. Notice that with this natural definition of sub-complexes, a small-weighted edge corresponds to an important edge, as it connects two neurons with close spatial proximity. In an induced graph $G(x,g)$, edge weight denotes information flow, not spatial proximity: a high-weighted edge thus corresponds to an important edge. To circumvent this issue, we replace the weight $w > 0$ with its opposite $-w$.

\subsection{Persistence Diagrams: Implementation Details}

In this paragraph, we give more detail about the different steps required to compute the persistent diagram for a given image $x$.

\paragraph{Step 1: Get the activations by layer.} As described in \cref{sec:induced_graph_threshold}, the induced graph depends both on the parameters of the networks and on the inner activations induced by $x$. Therefore, the first step is to perform a forward pass through our network and save all the intermediate activations (note that, in practice, we only focus on a subset of the layers as detailed in \cref{fig:hyperparams}). For layer $l$, we denote by $g_l(x) \in \mathbb{R}^{n_{l}}$ the inner activation.

\paragraph{Step 2: Matrices per layer.} To compute the induced graph, we need to weight the activations by the strength of the connection between neurons. For a linear layer parametrized by a weight matrix $W_l \in \mathbb{R}^{n_{l+1}\times n_{l}}$, this is straightforward and we can write:
\[ w_l = W_l g_l(x)\;\;.\]
For a convolutional layer, we need first to compute an equivalent weight matrix $W_l$ from the kernels $K_l$ (the "sparse fully connected counterpart").
When padding$=0$, stride$=1$ and nb\_channels$=1$, we can notice that the equivalent matrix is simply composed of Toeplitz matrices based on each row of $K_l$ stacked by block.
Here is an example.

$g_l(x)$ is the stacked version of $\begin{bmatrix} 
1 & 2 & 3\\
4 & 5 & 6 \\
7 & 8 & 9
\end{bmatrix}$ so that
$g_l(x) = \begin{bmatrix} 1 & 2 & 3 & 4 & 5 & 6 & 7 & 8 & 9
\end{bmatrix}^T$ and 
$K_l = \begin{bmatrix} 
10 & 20\\
30 & 40
\end{bmatrix}$. Then 
$$W_l = \begin{bmatrix}
10 & 20 & 0  & 30 & 40 & 0  & .  & .  & . \\
0  & 10 & 20 & 0  & 30 & 40 & .  & .  & . \\
.  & .  & .  & 10 & 20 & 0  & 30 & 40 & 0 \\
.  & .  & .  & 0  & 10 & 20 & 0  & 30  & 40
\end{bmatrix}$$ where the Toeplitz matrices are $T_1 = \begin{bmatrix} 
10 & 20 & 0\\
0  & 10 & 20
\end{bmatrix} \text{ and } T_2 = \begin{bmatrix} 
30 & 40 & 0\\
0  & 30 & 40
\end{bmatrix}$

The reasoning is similar in the general case where nb\_channels $\geq 1$, stride $\neq 1$ and padding $\geq 0$. In practice, we leverage the sparseness of these matrices when we build them and use the Numba package to accelerate the computations.

Note that the weight matrices per layer are computed once at the beginning of the process so that we can simply multiply $W_l$ and $g_l(x)$ to assemble the induced graph.

\paragraph{Step 3: Get the induced graph.} The induced graph is represented by its adjacency matrix $A \in \mathbb{R}^{n_1...n_L \times n_1...n_L }$. For NNs without any shortcuts (unlike ResNets for example), $A$ can be obtained by constructing a diagonal matrix by block, where the $l$-th block is simply the induced matrix of layer $l$.

\paragraph{Step 4: Edge selection.} We select the edges to keep based on the Magnitude Increase criteria as described in \cref{eq:threshold_param}: for each layer, we consider both $W_l$ and the initial weight matrix $W_l^{init}$ to compute the list of edges to be kept (independently of the activations). Then, for any input image $x$, we removed from its induced graph all the edges that are not in our list. As indicated in \cref{sec:induced_graph_threshold}, we chose to restrict ourselves to \emph{uniform selection parameter}, i.e. we keep the same fraction of edges $q$ in every selected layer.

\begin{figure*}
    \centering
    \subfigure[Trained NN]{\label{fig:trained_nn}\includegraphics[width=0.28\linewidth]{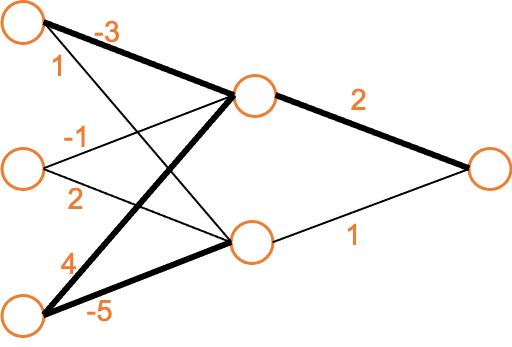}}
    \centering
    \subfigure[Induced graph]{\label{fig:induced_graph}\includegraphics[width=0.28\textwidth]{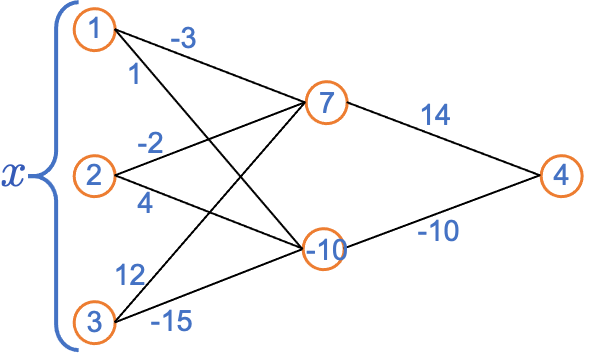}}
    \subfigure[Under-optimized induced graph]{\label{fig:thresholded_graph}\includegraphics[width=0.28\textwidth]{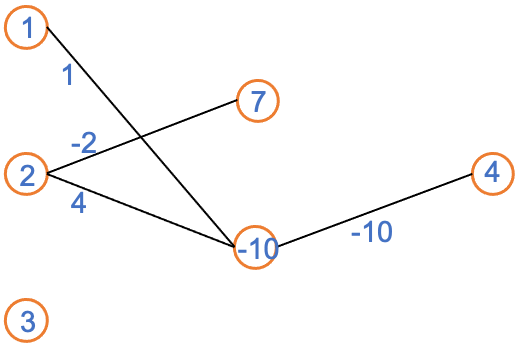}}
    
    \subfigure[Filtration. Corresponding persistent diagram: $\{(-5,\infty), (-3, \infty)\}$.]{\label{fig:filtration_graph}\includegraphics[width=0.95\textwidth]{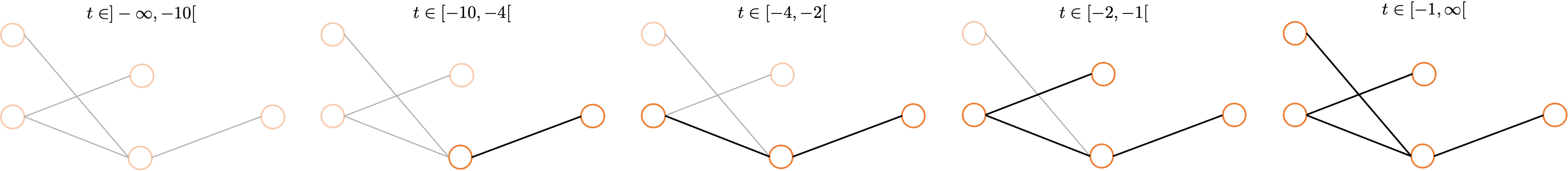}}
\caption{Persistence Diagram illustration - If we have a simple linear NN with its trained parameters in \cref{fig:trained_nn} (for simplicity, the initial values of the parameters were set to 0) and the selection parameter $q=0.5$, then: 1) we select only the \textit{thin} edges, not the thick ones, in \cref{fig:trained_nn}. 2) An example $x$ flows through the graph so that we obtain the corresponding induced graph in \cref{fig:induced_graph}. 3) Applying our selection parameter $q=0.5$, we restrain ourselves to the under-optimized induced graph in \cref{fig:thresholded_graph}. 4) The corresponding filtration is given by \cref{fig:filtration_graph}.}
\label{fig:the_dgm_pipeline}
\end{figure*}

\paragraph{Step 5: Compute the Persistent Diagram.} We use Dionysus \cite{dionysuslib} to compute the Persistent Diagram from a custom filtration where each edge $(u,v)$ appears at time $-|w^l_{u,v}|$ (strongest links appear first). An illustration of this process is given in \cref{fig:the_dgm_pipeline}. The persistence diagram we obtain is just a vector of tuples, containing the birth and death dates of every point in the persistence diagram.

\paragraph{Step 6: Computing the Sliced-Wasserstein gram matrices.} We can now compute the Sliced-Wasserstein kernel as proposed in \cite{carriere2017sliced}. The main parameter of the kernel is the number of sampled directions $M$: the higher $M$, the more accurate the value of the kernel. In our experiments, we set $M=50$, and tested that it was high enough to obtain a good approximation by comparing the results obtained with other values like $M=100$. For accelerating the computation of the Sliced-Wasserstein gram matrices, we use parallel C++ code.

\subsection{Algorithm}
\label{subsec:algo_appendix}
To compute PDs, we used the following simplified \cref{algo}. The complete code is available at \url{https://github.com/detecting-by-dissecting/detecting-by-dissecting}

\begin{algorithm}
\SetKwInOut{Input}{Input}
\SetKwInOut{Output}{Output}
\Input{a NN $g$ with parameters $W$ (after training) and $W^{init}$ (at initialization); a dataset $\mathcal{D}$; a parameter $q$; the SW kernel $K_{\dgm}$.}
\Output{An embedding dataset $\mathcal{F}=\{\Phi_{\dgm}(x,g) \, | \, \forall x \in \mathcal{D}\}$}
\For{each $x \in \mathcal{D}$}{
    \For{each pair of connected layers $(l,l')$}{
        \tcc{1 - Adjacency matrices}
        - Get $W_{l,l'}$ (parameter matrix) and $g_l(x)$ (output of layer $l$) as defined in \cref{sec:methods}\;
        - Compute $\forall \, i,j \; [A_{l,l'}(x)]_{i,j} = |[g_l(x)]_i * [W_{l,l'}]_{i,j} |$ \;
        \tcc{2 - Selecting under-optimized}
        \For{each matrix indexes $(i,j)$}{
        \uIf{$|[W_{l,l'}]_{i,j}| - |[W_{l,l'}^{init}]_{i,j}| \geq \text{quantile}(q)$}{
        $[A_{l,l'}(x)]_{i,j} \leftarrow 0$\;
        }
        }
    }
    \tcc{3 - Global adjacency matrix}
    Create $A(x)$ by stacking by block the $A_{l,l'}(x)$\;
    \tcc{4 - Persistence Diagram}
    - Compute $\Phi_{\dgm}(x,g) = \dgm(A(x))$\;
    - Add $\Phi_{\dgm}(x,g)$ to $\mathcal{F}$\;
}
\caption{Persistence Diagram embedding}
\label{algo}
\end{algorithm}


\section{What do PDs spot?}
\label{sec:how_pd_works}

\subsection{General illustration}


\begin{figure*}
    \centering
    \subfigure[Noisy circle 1]{\includegraphics[height=0.35\textwidth]{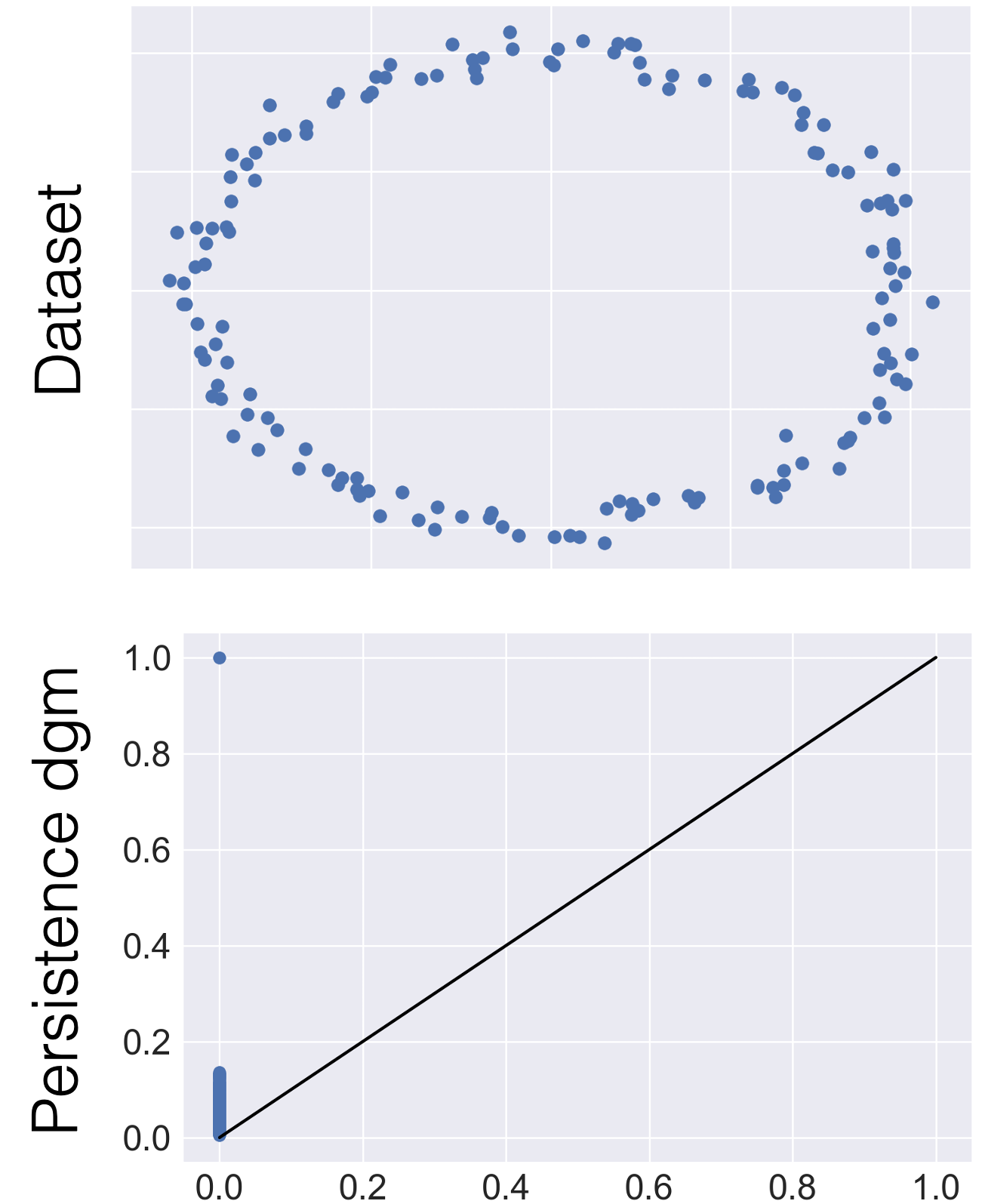}}
    \subfigure[Noisy circle 2]{\includegraphics[height=0.35\textwidth]{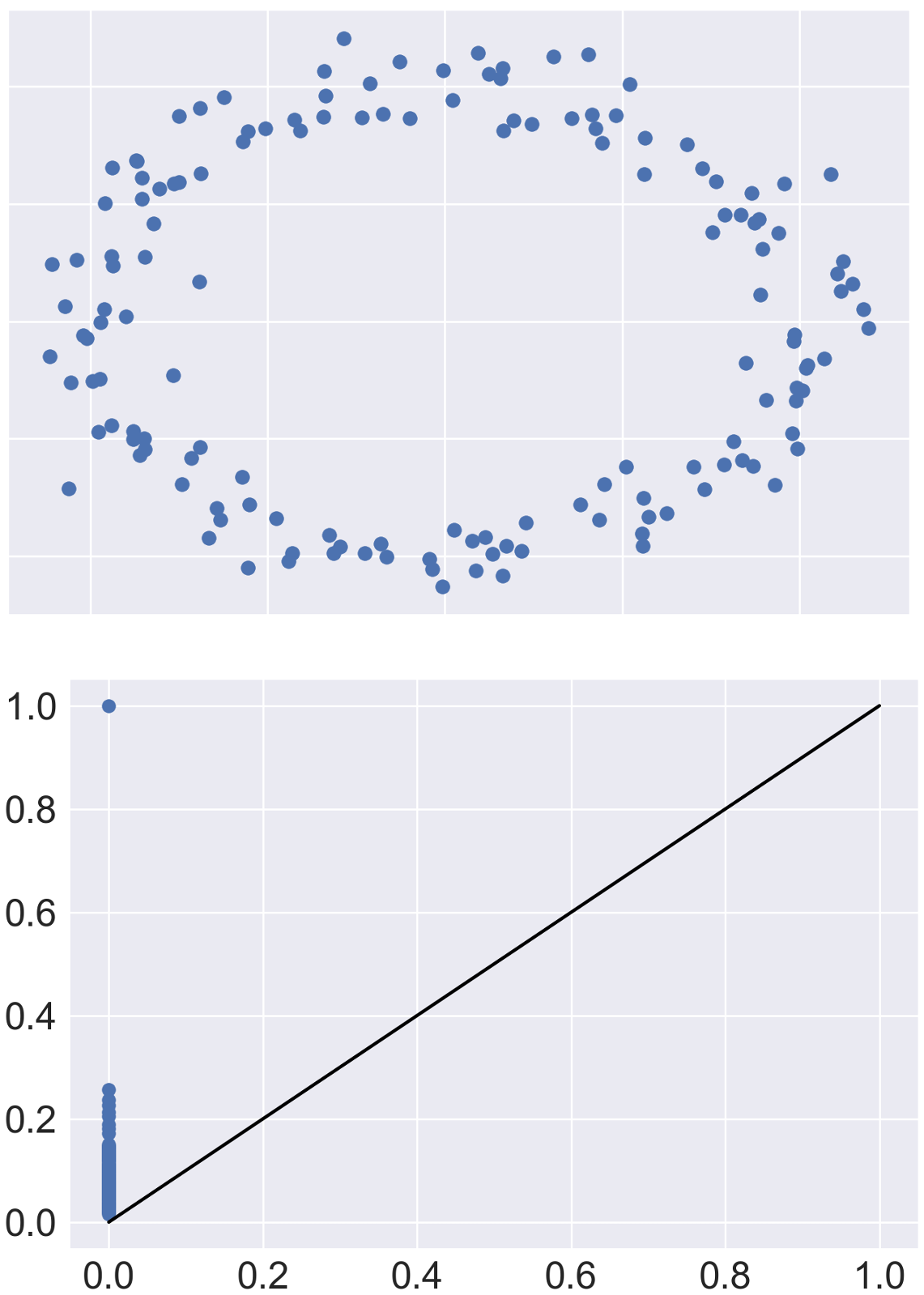}}
    \subfigure[Adv. circle]{\includegraphics[height=0.35\textwidth]{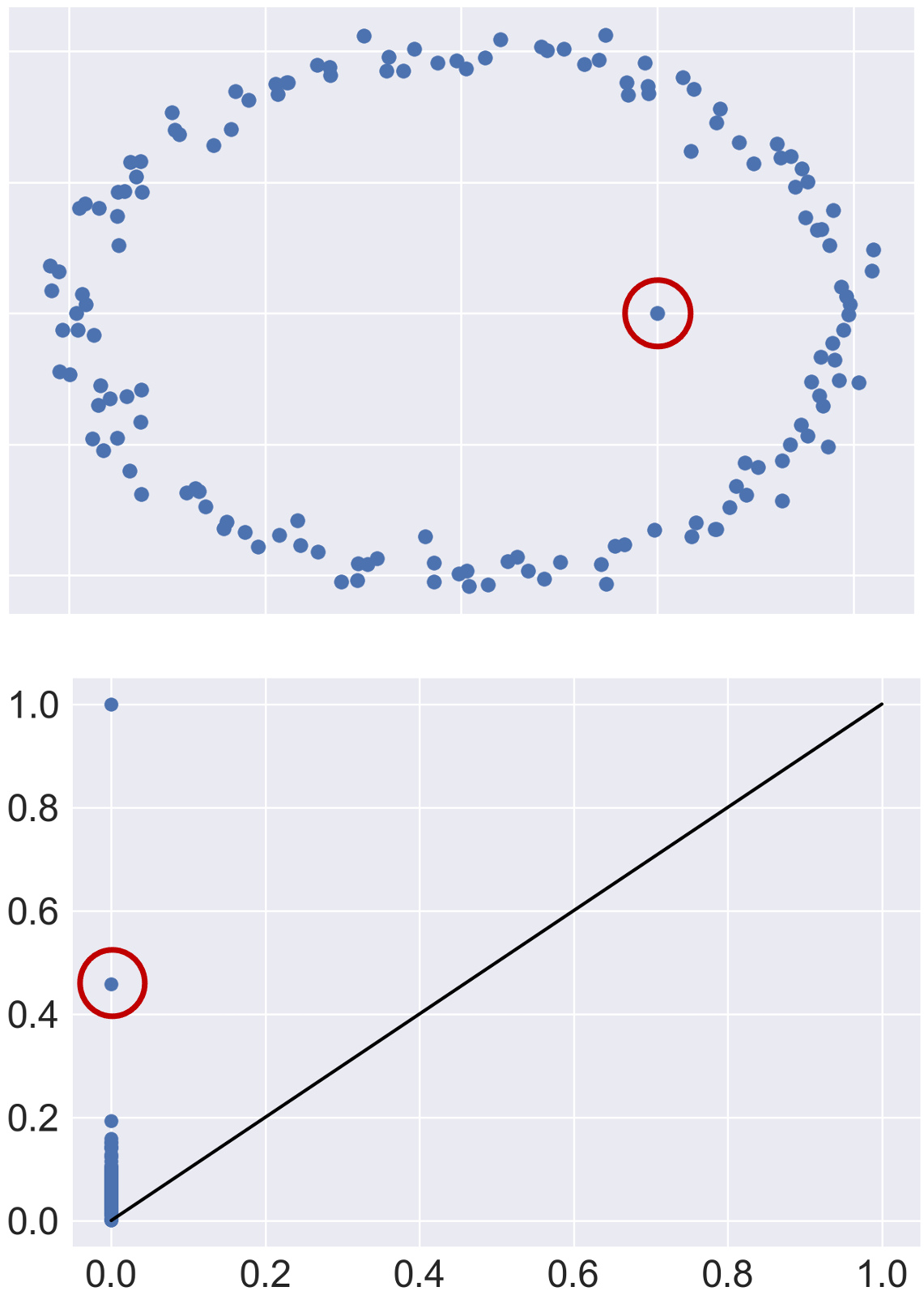}}
\caption{Persistence diagrams are stable to random noise, not to adversarial noise.}
\label{fig:dgm_illustration}
\end{figure*}

Persistence diagrams can identify structural properties of points clouds or graphs. In dimension 0, as previously stated, points in persistence diagrams represent the lifetime of \textit{holes}. An interesting property of persistence diagrams is that they are \textit{robust to noise}. It means that two noisy circles (the points in the dataset were generated following a circle equation to which a gaussian noise with mean$=0$ was added) will output very similar persistence diagrams. However, non-random noise, such as adversarial noise, can deeply modify the persistence diagram. We illustrate this feature in \cref{fig:dgm_illustration}. In the "adversarial" circle, we clearly see that even though there is only one adversarial point in the dataset, its position induce the presence of an abnormal point in the corresponding persistence diagram (emphasized with a red circle), whereas the two versions of the noisy circle dataset on the left output very similar diagrams.

\subsection{Adversarial robustness illustration}

\begin{figure*}
    \centering
    \subfigure[Toy dataset.]{\label{fig:toy_dataset}\includegraphics[height=0.25\textwidth]{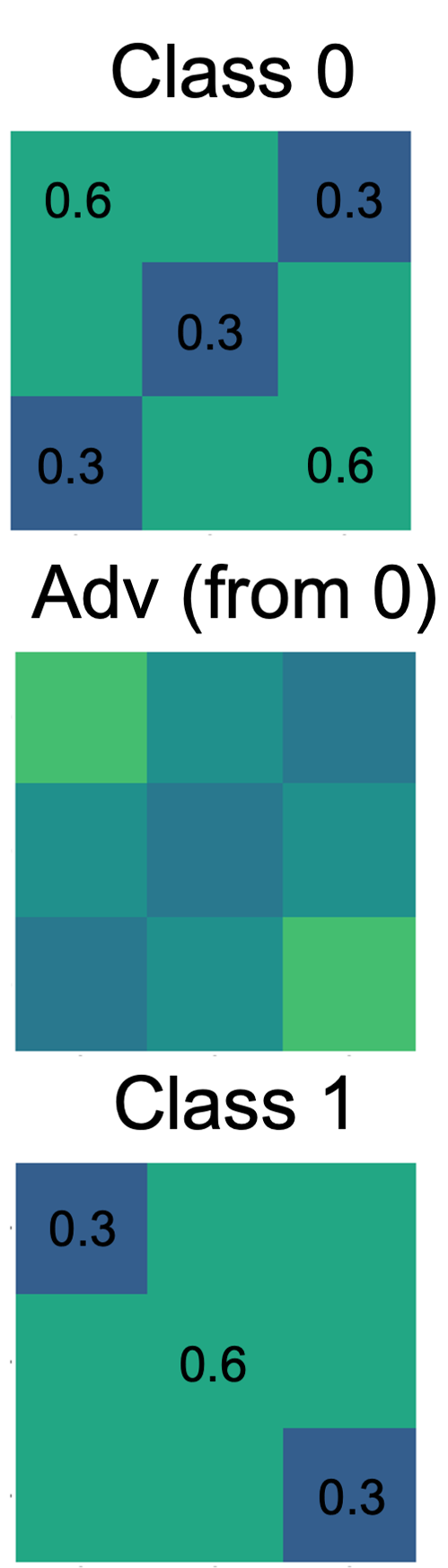}}
    \hspace{0.7cm}
    \subfigure[Dgms of four inputs illustrating transition phase]{\label{fig:toy_viz_dgms}\includegraphics[height=0.25\textwidth]{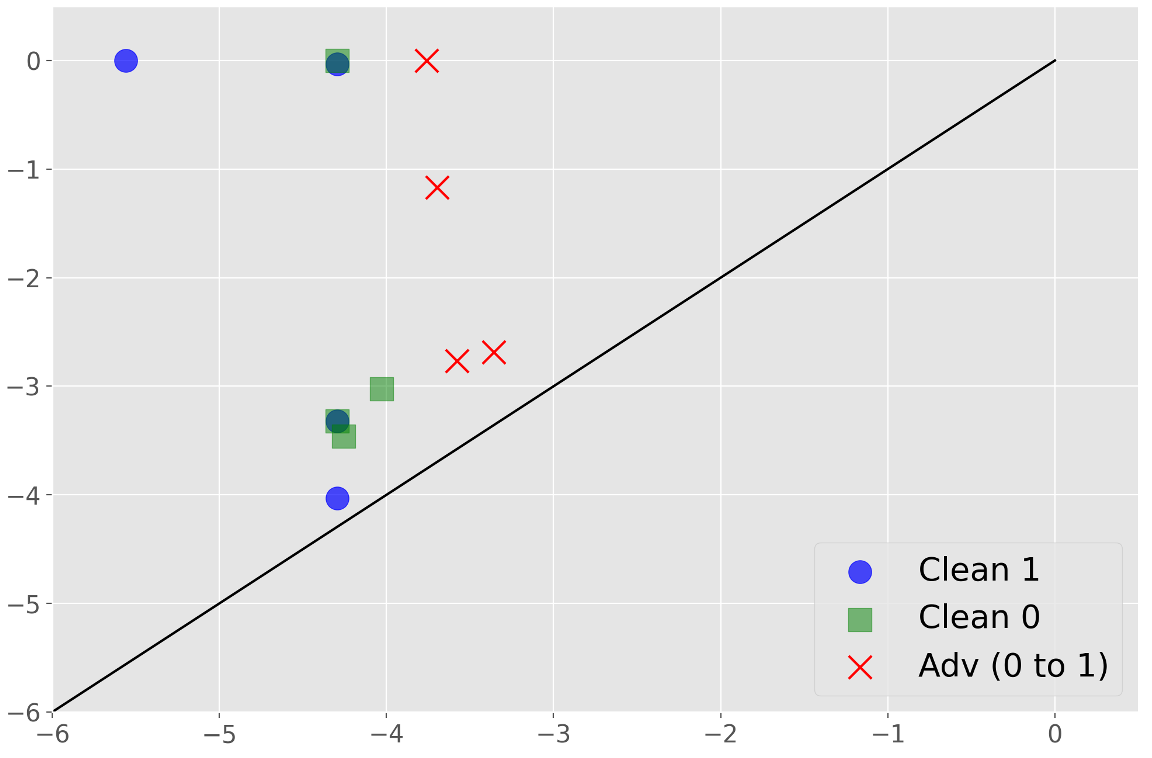}}
    \hspace{0.7cm}
    \subfigure[KDE plots of clean vs adv dgms]{\label{fig:toy_viz_kde}\includegraphics[height=0.25\textwidth]{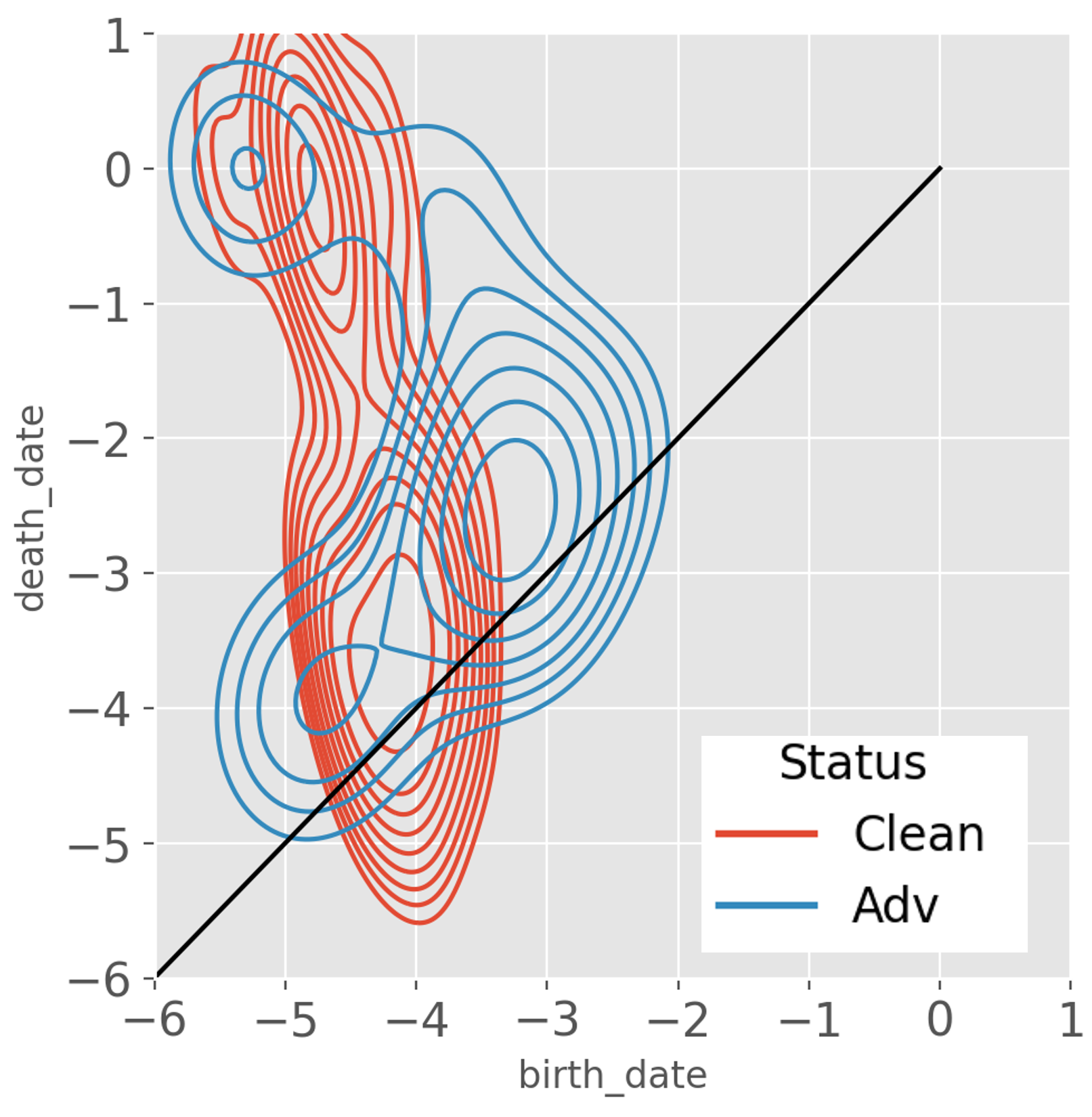}}
\caption{Persistence diagrams from clean vs adv inputs are highly dissimilar.}
\label{fig:toy_viz}
\end{figure*}

The robustness to noise property of persistence diagrams should result in having similar clean PDs (especially for inputs from the same class), but different from adversarial PDs because adversarial perturbations are non-random. Stemming from these non-random shifts in the structure of the induced graphs, we also expect a clear transition phase from the clean regime to the
adversarial one. Since PDs from classical tasks such as MNIST / LeNet have way too many points to be visually understandable, we trained a classical NN with one convolutional layer and two dense layers on a toy dataset. The dataset is a binary classification task on 3x3 images, where each pixel of an input conditionally to its class is drawn independently from a normal distribution with standard deviation$= 0.05$, and means as shown in \cref{fig:toy_dataset}. Our simple model outputs a standard accuracy of 0.99. Now, let us explore what PDs from clean vs adversarial inputs look like. We generated adversaries using PGD with $\varepsilon = 0.1$. In such a small setting, all PDs have very few points. However, even in this simple setting, we can illustrate that our hypotheses hold. 

\cref{fig:toy_viz_dgms} shows that PD from an adversary (created from a class 0 input, predicted as class 1) outputs a different behavior than the two clean ones: in addition to having larger birth dates, there is a particular point with a birth date and death date that do not correspond to any other point from either class 0 or class 1 diagrams. This behavior leads to a high distance between the adversarial diagrams and the clean diagrams from both classes. \cref{fig:toy_viz_kde} clearly shows that clean diagrams points lie in two very specific spots, whereas adversarial diagrams points are more dispersed, meaning that clean PDs (event from the two different classes) are quite similar, contrary to adversarial PDs.


\section{Related works}
\label{sec:related_works_appendix}

Our work is inspired by the preliminary works of Gebhart and Scharter \cite{gebhart2019adversarial}. However, our methodology overcome several limitations of their work and differ in many aspects. We summarize here the outline of their methods, before discussing the limitations and the differences with our paper.

\paragraph{Gebhart and Schrater methodology.} In \cite{gebhart2019adversarial}, the authors also use topology tools, and more specifically persistence diagrams, to detect adversarial examples. Their work do not rely on under-optimized edge, which is at the core of our work here. The simply compute a persistence diagrams on a whole induced graph, select some points of the persistence diagram and reconstruct a sub-graph based on them. Then, they perform an analysis of the said sub-graph (e.g. eigenvalues of the Laplacian matrix).

\paragraph{Limitations.} Two major limitations from \cite{gebhart2019adversarial} are:
1) Uninterpretable results: they detect differences in the topology of clean vs adversarial induced graphs, but are not able to provide an explanation stating why such differences are visible. On the contrary, in our work, we first provide an hypothesis about how adversarial examples operates, and verify this hypothesis thanks to topological tools. Our work is then aligned with the objective of improving our understanding of adversarial examples.
2) Scalability: computing a persistence diagram depends on the number of edges and neurons in the graph, which is very large even for quite small NNs like LeNets. As \cite{gebhart2019adversarial} compute persistence diagrams for each input on the entire NN, the computation complexity is much too high to study larger networks, and indeed, the experiments focus on 3 or 4- layers CNNs. Their method does not apply to larger networks. On the contrary, by selecting only under-optimized edges in the induced graph before computing the persistence diagram, our PD method is scalable.

A third limitation is the complexity of their methods. Reconstructing a sub-graph is not straightforward, and extracting relevant features directly from graph objects is, again, not straightforward. Many possibilities can be explored: computing a classical metric for graph (the eigenvalue of the Laplacian matrix is only one of them), using custom features (the vectorization method used in \cite{gebhart2019adversarial} is only one of them too), using GNNs, etc. Finding the most relevant metric is challenging, which is not the case when studying directly the persistence diagram as we do.

\paragraph{Fundamental differences with our work.} 

The main difference of approach between \cite{gebhart2019adversarial} and our work is that, when studying different inputs (clean or adversarial), we study the same edges for all. The persistence diagrams corresponding to these inputs are different (because the weights on these edges are different), but the structural object remains the same. On the contrary, \cite{gebhart2019adversarial} study input-specific edges. Both approaches are relevant and interesting, however, our approach enables us to provide more insights on these specific edges and on the behavior of adversarial inputs. This corresponds to our analysis of under-optimized edges.


\section{Experiments details}
\label{sec:exp_details_param}

\subsection{Architectures used in the experiments}

\begin{figure*}
    \centering
    \subfigure[MNIST and Fashion MNIST LeNet architecture]{\includegraphics[width=0.98\linewidth]{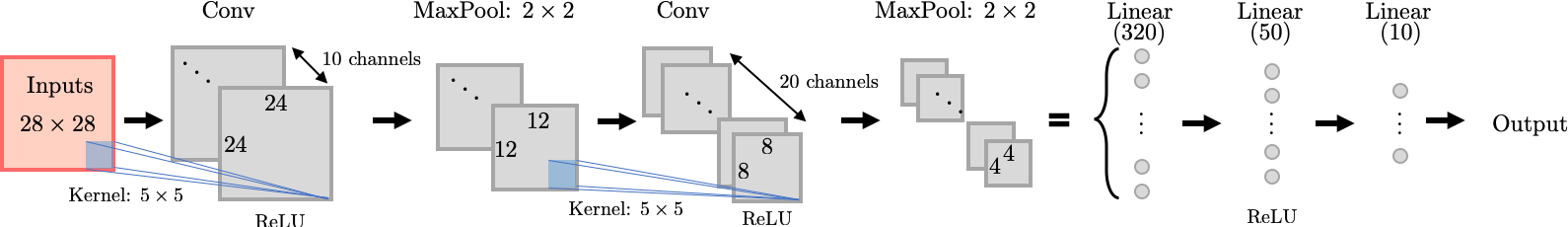}}
    
    \centering
    \subfigure[SVHN and CIFAR10 ResNet 18 architecture]{\includegraphics[width=0.98\textwidth]{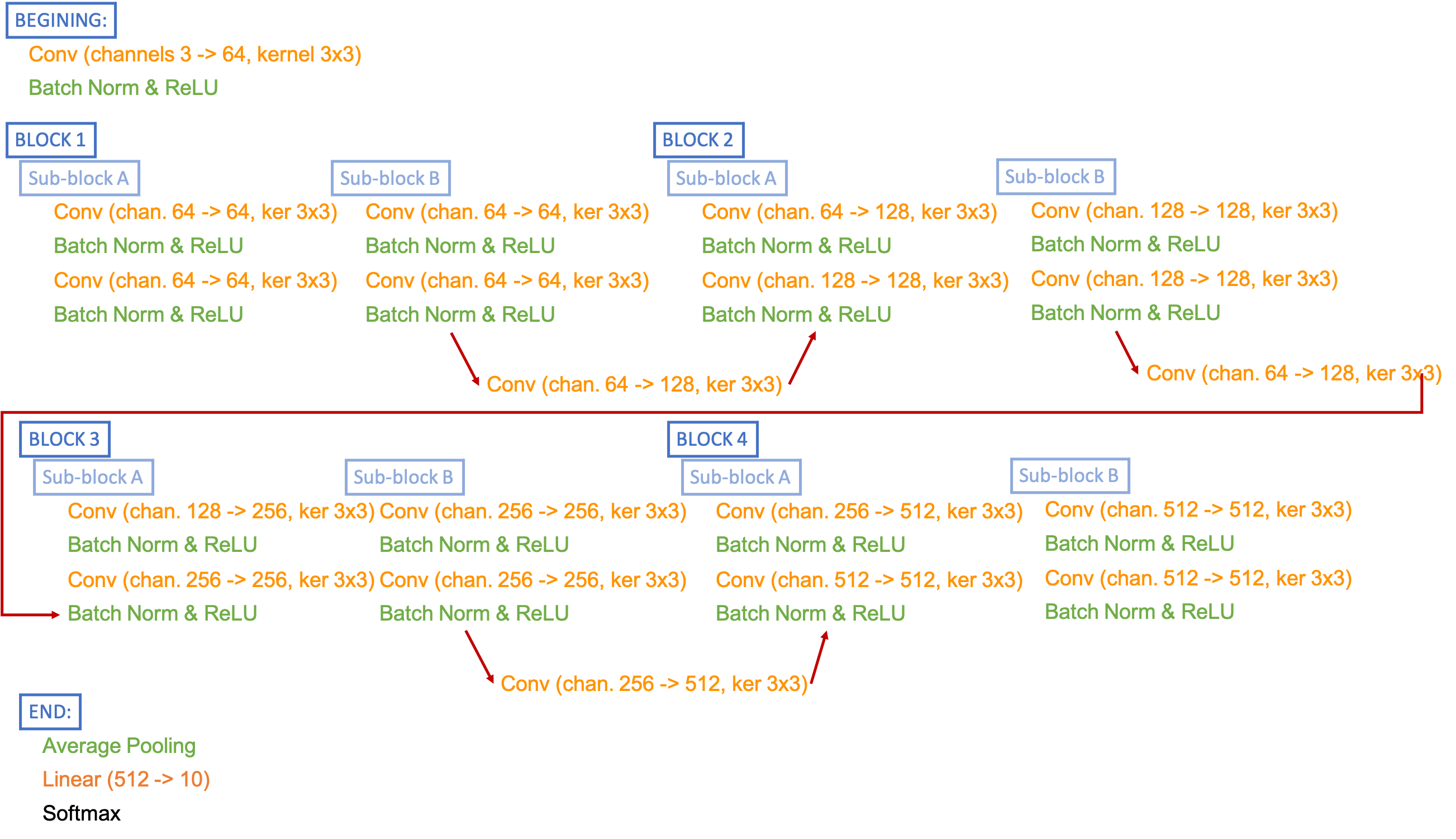}}
\caption{Architectures used in the paper.}
\label{fig:archi}
\end{figure*}

The two LeNets and the two ResNets used are represented in \cref{fig:archi}.

\subsection{More details on time complexity.}

\begin{figure}[h!]
    \centering
    \includegraphics[width=0.5\textwidth]{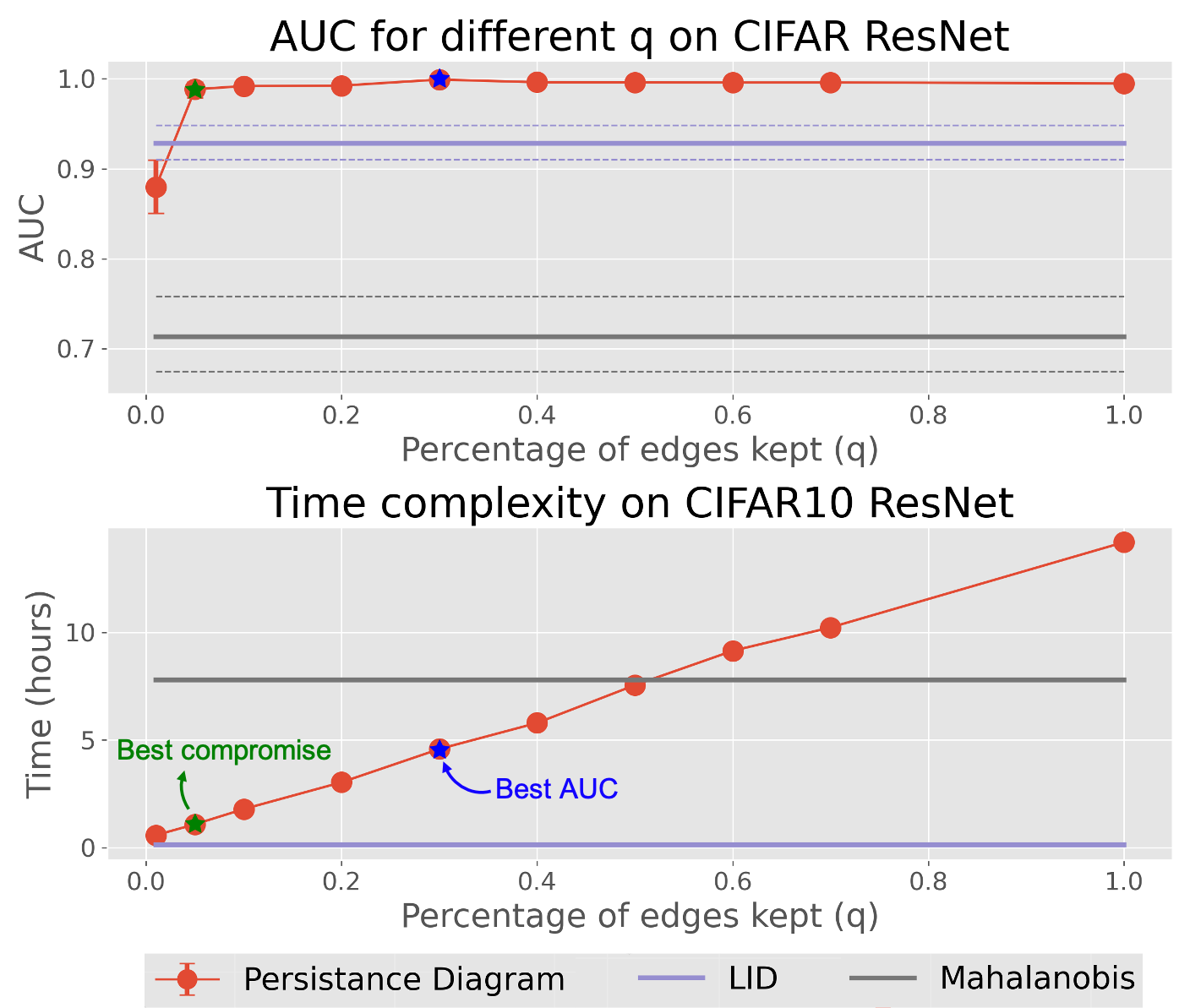}
    \caption{Detection AUC (up) and time (down) as a function of $q$ (CIFAR10 ResNet vs PGD $\varepsilon=0.05$).}
    \label{fig:time_complexity}
\end{figure}

\cref{fig:time_complexity} illustrates the fact that the time complexity of our PD methods grows linearly with parameter $q$. However, one can see that even small values of $q$ yield great detection results, with almost no compromise on the AUC (green star). Note that Mahalanobis requires the estimation of large precision matrices (one for each considered layer, of size nb neurons x nb neurons), which makes it substantially slower than LID.

\subsection{Training details.} The usual procedure was used for training, by separating the datasets into training, validation, and test sets and using an Adam optimizer (for LeNets) and an SGD optimizer (for ResNets). The learning rate was set to 0.001 for the LeNets, and a one-cycle policy (see \cite{smith2017cyclical}) with varying learning rates in the range $[0.008,0.12]$ for SVHN and CIFAR10 ResNets. The number of epochs was set to $50$ for MNIST LeNet and $100$ for the others.

Note that the ResNet32 model used for CIFAR100 was a pre-trained model without further training, downloadable here: https://github.com/chenyaofo/pytorch-cifar-models/releases/download/resnet

We ran all our experiments on a computer equipped with 1 GPU (Tesla V100-PCIE-16GB) and 60Gb of RAM.

Our code is available here: \url{https://github.com/detecting-by-dissecting/detecting-by-dissecting}

\subsection{Attacks details.}
\label{sec:attacks_details}
Recall that PGD attack~\citep{kurakin2016adversarial} is defined by: $x^{adv}_0 = x$ and $x^{adv}_{t+1} = \text{Clip}_{x, \varepsilon} \left( x^{adv}_{t} + \varepsilon_{\text{iter}} \, \text{sign}\left( \Delta_x  \text{L}(\theta, x, y) \right) \right).$ for each $t \in [\![1,T]\!]$. In our experiments, we set $T=50$ and $\varepsilon_{iter}=2*\varepsilon/50$ and different $\varepsilon$ values (reported in the results).

The objective of CW~\citep{carlini2017towards} is to find $\delta^* = \text{arg}\text{min}_{\delta} ||\delta||_{2} + c f(x+\delta)$ with $f$ a well-chosen function. In our experiments, we set the number of binary search steps to find $c$ to 15; the number of iterations to optimize the objective function to 50 (Adam optimizer).

\subsection{Experimental pipeline.}
There are 3 steps in the detection pipeline:
\emph{1)~Pre-processing.} We create first a (successful) adversarial dataset by running an attack on the NN and clean inputs. For the clean dataset, we keep only examples that were not involved in the creation of the adversarial dataset.
\emph{2)~Feature extraction.} We apply our methods (or SOTA) to the clean and adversarial datasets (see \cref{algo} in \cref{subsec:algo_appendix} for PD).
\emph{3)~Detector.} An SVM is trained with the features of each method, and its outputs enable us to compute any detection metric (namely the AUC).

Moreover we ran \textit{unsupervised} and \textit{supervised} experiments. Supervised ones use adversarial data during training: by assuming something about the type of attack, they are uninformative about the generalization ability of the method (they give a false sense of security).
The unsupervised experiments are using a one-class SVM trained only on clean data: it is a better setting to evaluate detection methods. We only show unsupervised results in the main paper (see  \cref{sec:supervised_results} for the rest, where our method still outperforms SOTA methods). Note then that SOTA results are not as high in this setting compared to the results reported in other papers.

\subsection{Computing the AUC.} \label{sec:computing_auc} As a reminder, when computing the AUC, the attack method (and the attack strength) and the detection parameters (like the parameter $q$ for PD and RG) are given. To compute this score, the SVM needs to have a kernel as input. For the PD method, the kernel used was the Sliced-Wasserstein kernel. For the three other methods (RG, LID and Mahalanobis), the kernel used was just the classical \textit{Radial Basis Function} (RBF) kernel, defined as:

\begin{eqnarray} K_\Phi(x, x') = \exp\left(-\frac{1}{2\sigma^2} \Vert \Phi(x) - \Phi(x') \Vert^2  \right),
\end{eqnarray}
where $\Phi$ denotes the features for each method, e.g. $\Phi_{\text{RG}}(x):= \Phi_{\text{RG}}(x,g) = \text{Vect}(W^q(x,g))$, where $W^q(x,g)$ is the matrix of weights of the under-optimized induced graph $G^q(x,g)$.

SVM outputs scores for each input: if it is above a discrimination threshold, the input is flagged as clean (otherwise, flagged as adversarial). The ROC curve is a plot representing the TPR as a function of the FPR when the discrimination threshold varies. The AUC is the integral of the ROC function (so that the discrimination threshold is integrated out), and represents how well the detector can separate the two classes (the higher the AUC, the better).

\paragraph{Confidence Intervals}
\label{sec:confidence_intervals}

The main source of variability of a run comes directly from the variability of the dataset. For a fixed detector, we denote by $F$ the distribution of the images. We want $[p, q]$ that satisfies  (80\%-confidence interval)

\begin{eqnarray*}
\mathbb{P}_F\left\lbrace AUC < q \right\rbrace = 0.1 \text{  and  }\mathbb{P}_F\left\lbrace AUC > p \right\rbrace = 0.1
\end{eqnarray*}

To estimate $[p, q]$, we use resampling and estimate the AUC on $100$ bootstraps of size $n//2$ (where $n$ is the total number of samples). It can be shown (see for instance \cite{johnson2001introduction}) that a good approximation of $[p, q]$ is given by 
\begin{eqnarray*}
[2 \hat{AUC} -c_{90}, 2 \hat{AUC}-c_{10}]\;\;,
\end{eqnarray*}

where $\hat{AUC}$ is the AUC estimated on the $n$ samples, $c_{10}$ (resp. $c_{90}$) is the 10-th percentile (resp. 90-th percentile) of the $100$ bootstrapped AUCs.

\section{Hyperparameters for methods used in the paper}
\label{sec:hyperparameters}

We cross-validated the parameter values for all parameters presented below, and kept only the best ones that were used afterward in our experiments.

\subsection{Selection parameter for PD and RG methods.}

Recall that the parameter used for our PD and RG methods is denoted by $q$: it is the proportion of edges kept for the construction of the induced graph. We use the same value $q$ for selected layers (uniform selection), thus we have to identify the layers kept in the analysis, and then find the paramater to use for all these layers. Note that the parameter was optimized on the PD method, and kept the same for the RG method. 

\begin{figure}[ht!]
    \centering
    \begin{tabular}{|c|c|c|}
        \hline
         Models & Max percentile $q$ & List of layers \\
         \hline
         \hline
         MNIST LeNet & 0.025 & All layers\\
         Fashion MNIST Lenet & 0.05 & All layers\\
         SVHN ResNet & 0.275 & Last conv. and linear layers\\
         CIFAR10 ResNet & 0.3 & Last conv. and linear layers\\
         \hline
    \end{tabular}
    \caption{Selection parameter used for PD and RG methods in the experiments}
    \label{fig:hyperparams}
\end{figure}

\subsection{Hyperparameters for the LID method.} LID has two parameters that we cross-validated.

\begin{figure}[ht!]
    \centering
    \begin{tabular}{|c|c|c|}
        \hline
         Models & Nearest Neigh. \% & Batch size \\
         \hline
         \hline
         MNIST LeNet & 0.08 & 250 \\
         Fashion MNIST Lenet & 0.02 & 250 \\
         SVHN ResNet & 0.05 & 150 \\
         CIFAR10 ResNet & 0.1 & 50 \\
         \hline
    \end{tabular}
    \caption{LID parameters used in the experiments}
\end{figure}

\subsection{Hyperparameters for Mahalanobis method.} Mahalanobis has two parameters: the first one, $\epsilon_\text{preprocessing}$, controls the size of the noise added to the input, in order to make in- and out-of-distribution samples more separable. We set this parameter to 0.0.
The second one is the layer selected for the analysis. When it was available (for the two setups using ResNet), we used the same layers as the one used by the authors of Mahalanobis in \cite{lee2018simple}. For the experiments using LeNet, we kept the last two linear layers.

\begin{figure}[ht!]
    \centering
    \begin{tabular}{|c|c|}
        \hline
         Models & Selected leyers \\
         \hline
         \hline
         MNIST LeNet & Last two linear layers \\
         Fashion MNIST Lenet & Last two linear layers \\
         SVHN ResNet & Last layer of each four ResNet block  \\
         CIFAR10 ResNet & Last layer of each four ResNet block \\
         \hline
    \end{tabular}
    \caption{Mahalanobis parameters used in the experiments}
\end{figure}

In addition, note a substantial difference between our experiments and theirs when evaluating against PGD attack: the $\varepsilon$ parameter in \cite{lee2018simple}'s implementation corresponds in fact to $\varepsilon_{iter}$ in our paper: thus, at the end, when they run a PDG attack with strength $\varepsilon$, the resulting perturbation is much higher, of size $\varepsilon /
\times \text{ number of iteration for PGD}$. This leads to better detection results since they evaluate on much stronger attacks.


\section{Additional detection results}
\label{sec:additional_results}

\subsection{Supervised Results}
\label{sec:supervised_results}

As mentioned before, supervised results can give a false sense of security because, in practice, one cannot anticipate which algorithm will be used to craft an adversarial example (see \cref{tab:unsup_vs_sup_v2}): for LID and Mahalanobis, the supervised AUCs are noticeably better than the unsupervised ones, with confidence intervals for these almost not overlapping; on the contrary, PD is more stable between these settings (the difference is around six times smaller). We report results from this unsupervised setting. To compare with literature (where most of the results are reported under the supervised setting) we also provide supervised results in the Appendix. Keep in mind that great results on supervised experiments are easier to achieve than on unsupervised experiments because, obviously, the task is harder.

\begin{figure}[h]
    \centering
    \begin{tabular}{|c||c|c|c|}
        \hline
         & Sup. & Unsup. & Diff \\
         \hline
         \hline
         PD & 0.884 {\small[0.858, 0.910]} & 0.873 {\small[0.851,0.902]} & \textbf{0.011} \\
         LID & 0.835 {\small[0.799, 0.870]} &  0.776 {\small[0.744, 0.817]} & 0.059 \\
         Maha & 0.772 {\small[0.737, 0.811]} & 0.712 {\small[0.664, 0.748]} & 0.06 \\
         \hline
    \end{tabular}
    \caption{Supervised vs unsupervised detection of adversarial examples. Showing AUC for ResNet / SVHN subject to PGD attacks with $\varepsilon = 0.01$. Smaller diff. is better.}
    \label{tab:unsup_vs_sup_v2}
\end{figure}

However, results using the \textit{supervised} setting are quite similar to those obtained under the \textit{unsupervised} setting (the AUC are overall higher, because the task is simpler): the hierarchy between the detection methods is identical, with Persistence Diagram providing the best results, followed by LID and Mahalanobis. Note that, as mentioned in the main paper, some AUC results are significantly higher in the supervised setting (Raw Graph, Mahalanobis, etc.), illustrating the false sense of security we can get by studying only supervised results.

\begin{figure*}
    \centering
    \subfigure{\includegraphics[width=0.7\linewidth]{fig/legend_exp_plot.png}}
    \setcounter{subfigure}{0}
    
    \centering
    \subfigure[LeNet / MNIST]{\includegraphics[width=0.245\textwidth]{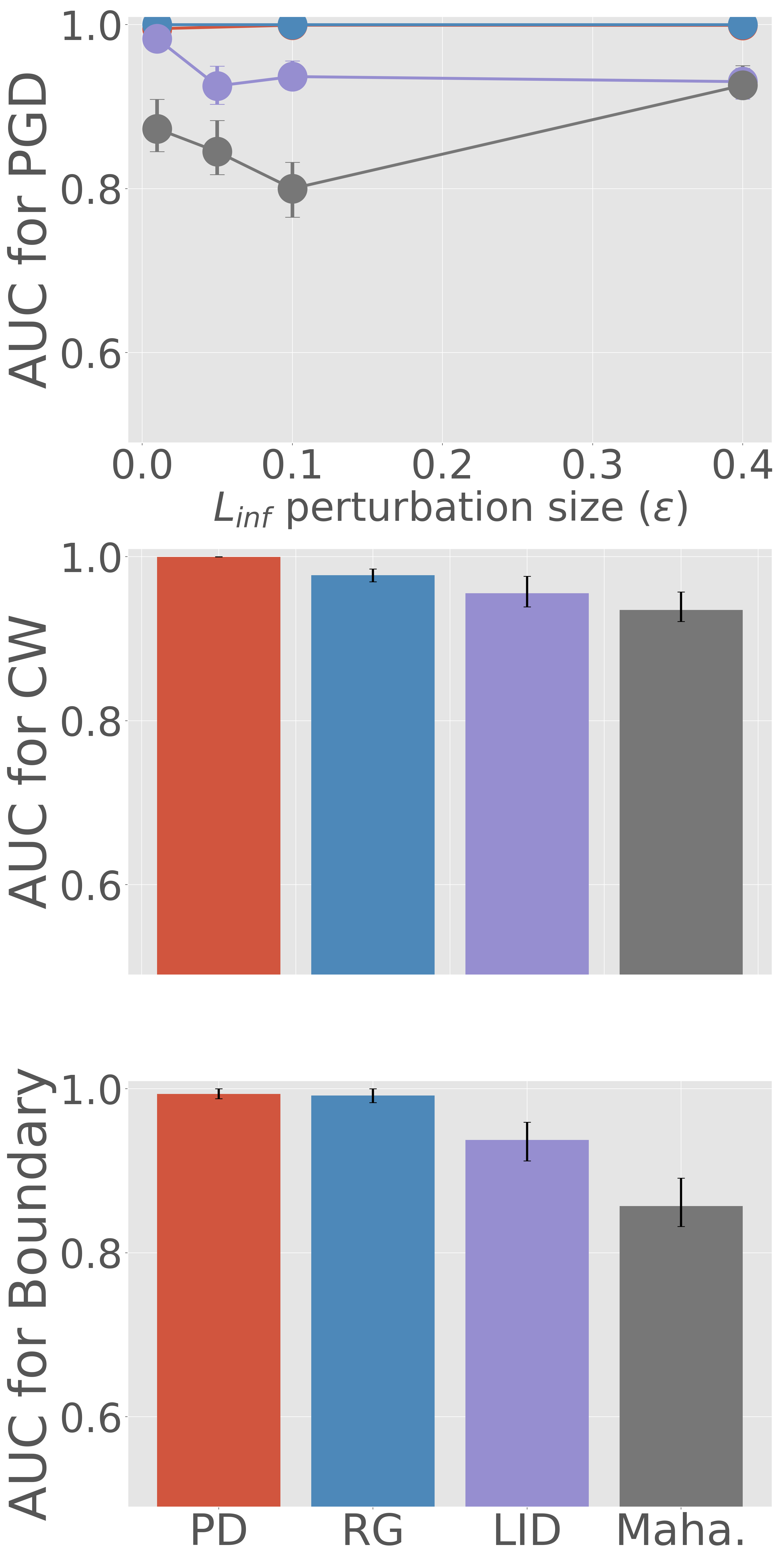}}
    \subfigure[LeNet / Fashion MNIST]{\includegraphics[width=0.245\textwidth]{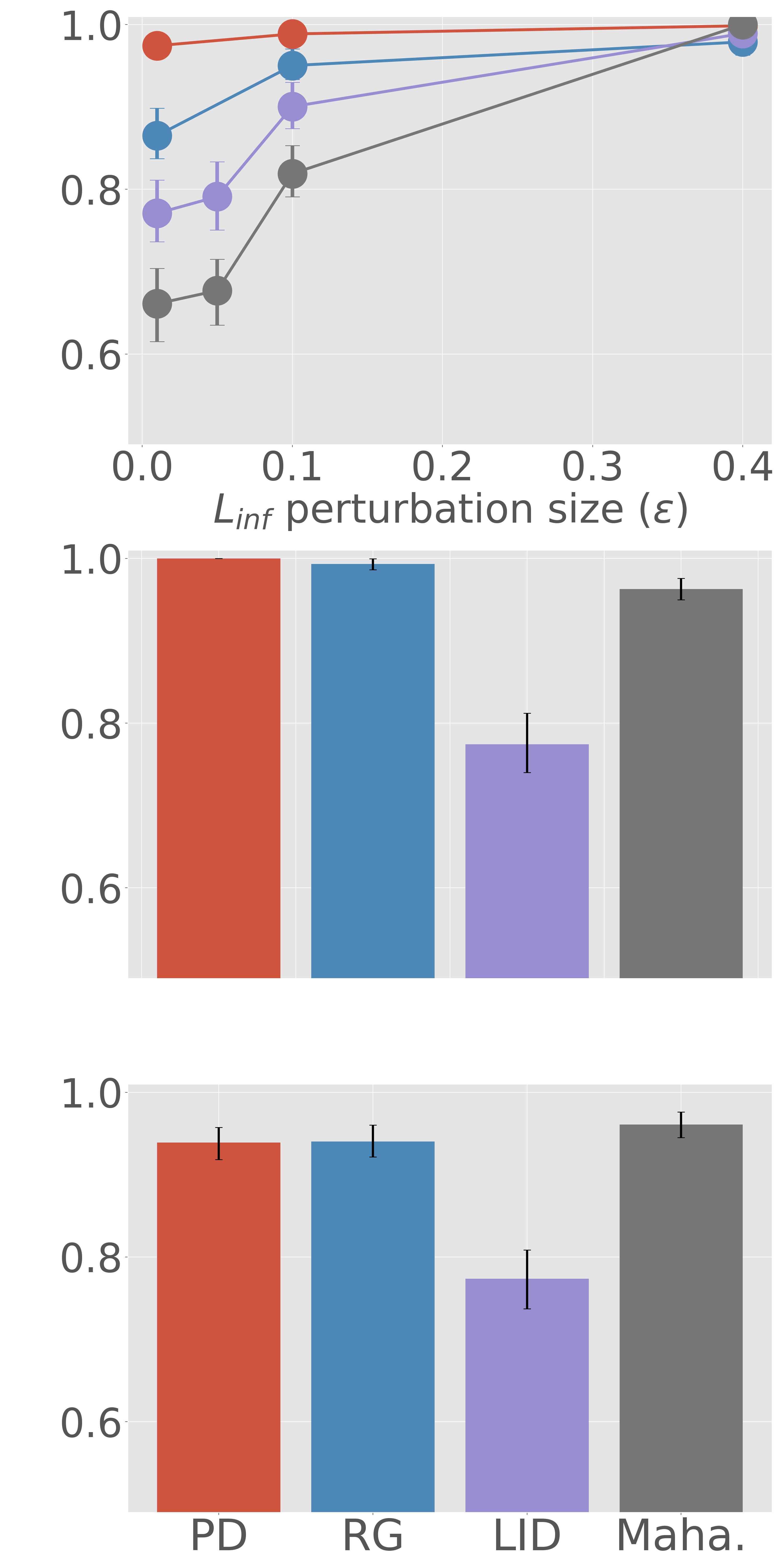}}
    \subfigure[ResNet / SVHN]{\includegraphics[width=0.245\textwidth]{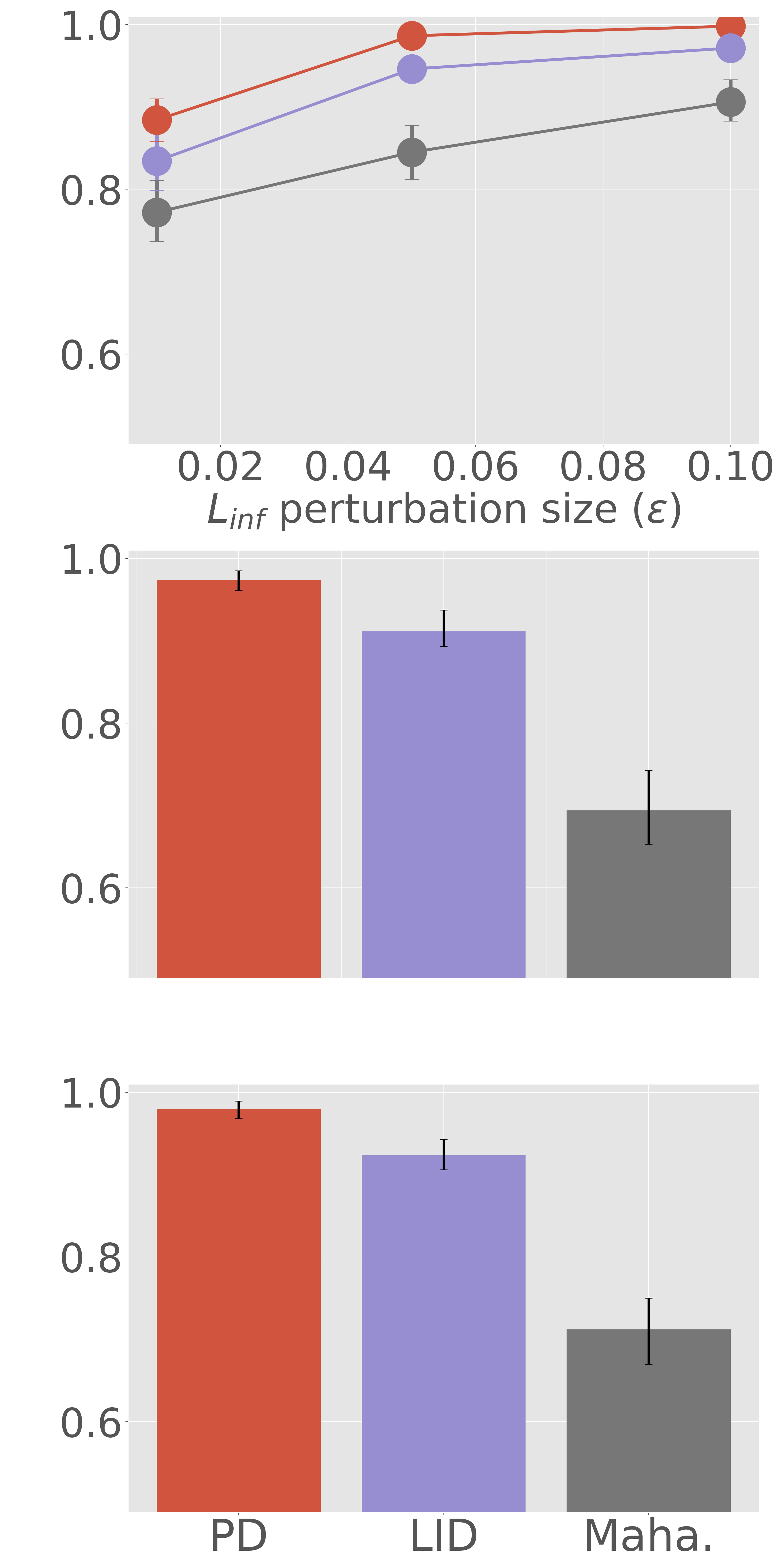}}
    \subfigure[ResNet / CIFAR10]{\includegraphics[width=0.245\textwidth]{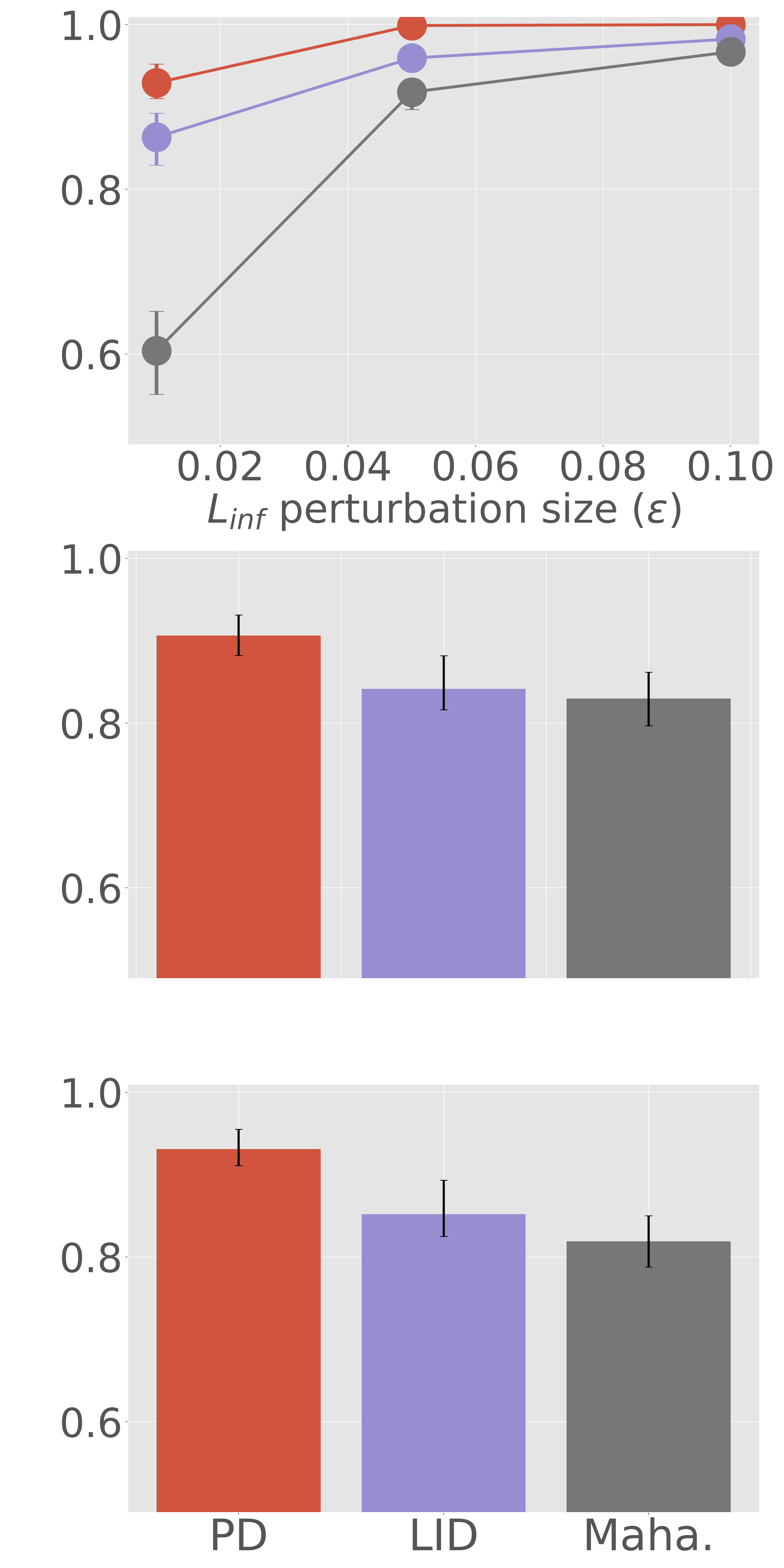}}
\caption{Supervised results - Showing detection AUC for different detection methods (legend) against different kinds of adversarial attacks (rows) and model architectures and datasets (columns)}
\label{fig:supervised_exp}
\end{figure*}


\subsection{Unsupervised results on transferred attacks}
\label{sec:unsup_transf_results}

\begin{figure*}
    \centering
    \subfigure{\includegraphics[width=0.7\linewidth]{fig/legend_exp_plot.png}}
    \setcounter{subfigure}{0}
    
    \centering
    \subfigure[LeNet / MNIST]{\includegraphics[width=0.25\textwidth]{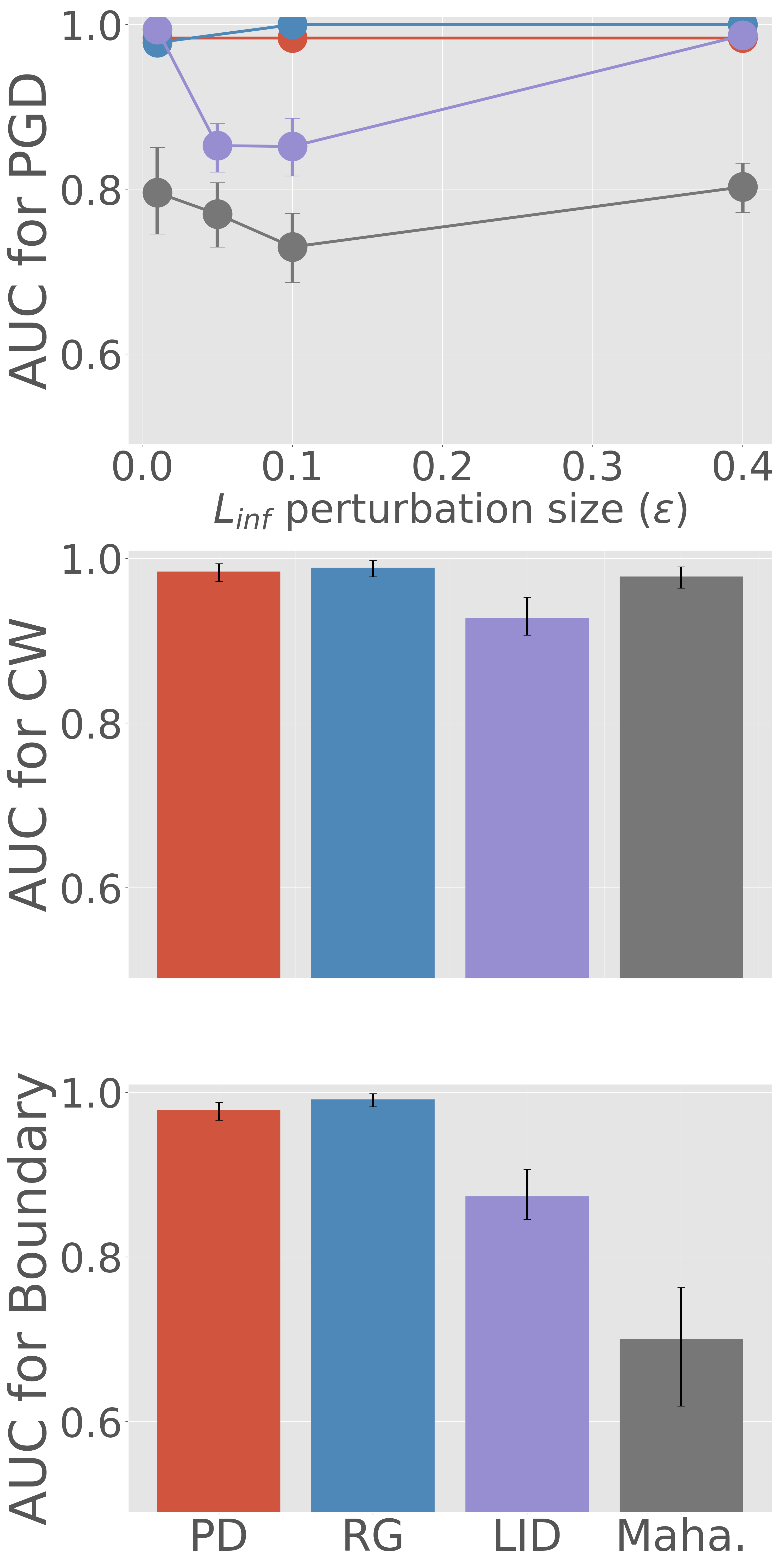}}
    \subfigure[LeNet / Fashion MNIST]{\includegraphics[width=0.25\textwidth]{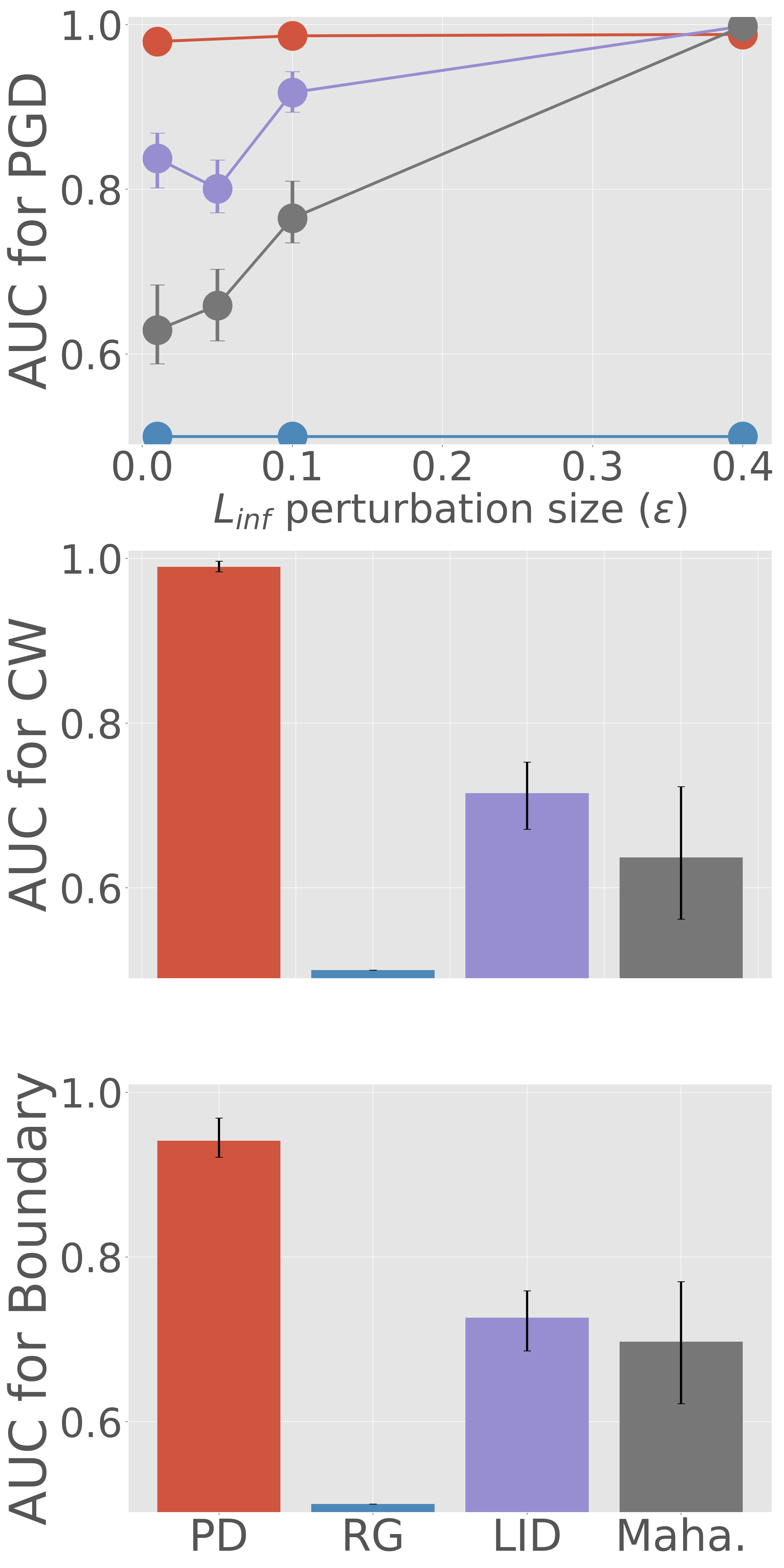}}
\caption{Transferred attacks results - Detection AUC for different detection methods (legend) against different kinds of adversarial attacks (rows) and model architectures and datasets (columns).}
\label{fig:trsf_exp}
\end{figure*}

We also ran experiments using transferred attacks on MNIST and Fashion MNIST LeNets, reported in \cref{fig:trsf_exp}. Transferred attacks were generated on control models (using the same LeNet architecture), and successful adversaries on these control models were saved. Then, these attacks were submitted to our original target models, and detection methods were launched to flag these adversaries. The results reported here correspond to a black-box setting.

The results are quite similar to those observed for the white-box setting, with our PD method still better than LID and Mahalanobis. As mentioned in \cref{sec:preliminary_exp}, the three main methods (PD, LID, Mahalanobis) seem to generalize well in this black-box setting.

\subsection{Unsupervised results on CIFAR100}
\label{sec:unsup_cifar100_results}

\begin{figure}
    \begin{minipage}{\linewidth}
    \begin{center}
    \includegraphics[width=0.30\linewidth]{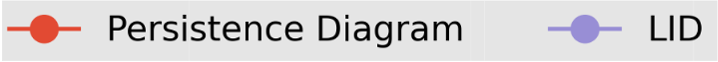}
    \par\vfill
    \includegraphics[width=0.30\linewidth]{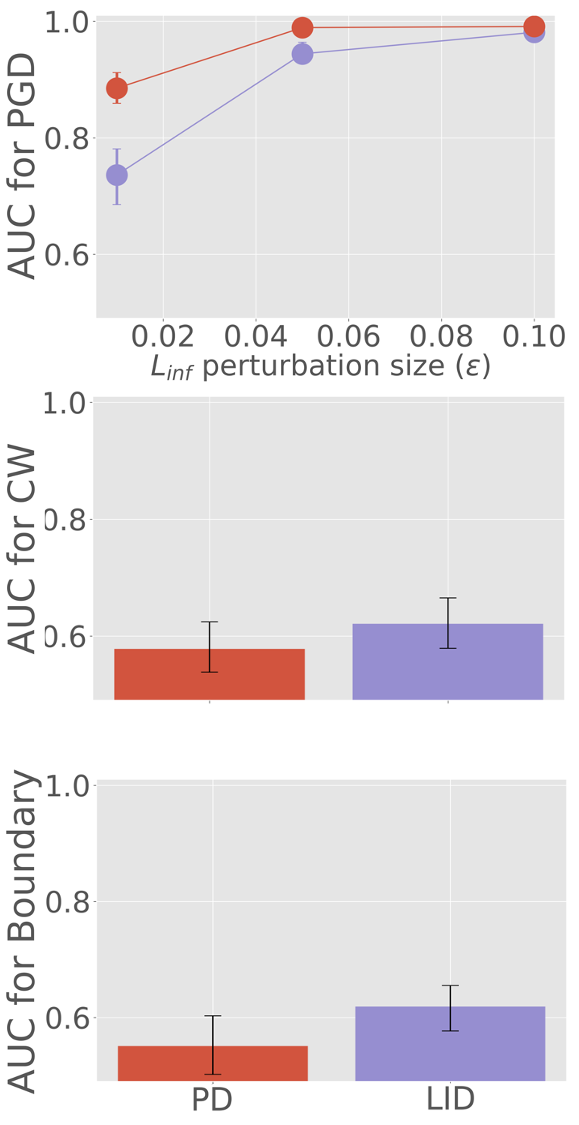}
    \label{fig:cifar100_}
    \end{center}
\end{minipage}
\caption{Results for ResNet32 / CIFAR100 (unsupervised).}\label{fig:cifar100_exp}
\end{figure}

To experiment with a higher number of classes, we consider the CIFAR100 dataset. This dataset is similar to CIFAR10, except it has 100 classes containing 600 images each. We consider a ResNet32 pretrained on this dataset (from  \url{https://github.com/chenyaofo/pytorch-cifar-models/}). We provide in \cref{fig:cifar100_exp} the detection results. Note that for this experiment, we don't have access to the initial weights of the model and therefore, we cannot identify the under-optimized edges as presented in \cref{sec:induced_graph_threshold}. We replace in \cref{eq:threshold_param} the \emph{Magnitude Increase} criteria $|(W_l)_{u,v}| - |(W_l^{init})_{u,v}|$ by the simpler \emph{Large Final} criteria $|(W_l)_{u,v}|$ also studied in \cite{frankle2018lottery,zhou2019deconstructing}. 

This task being significantly harder than the ones studied before (100 vs 10 classes), adversaries were expected not only to be harder to detect but also to behave differently according to the adversarial attack used: we thus decided to adapt our parameters to each attack. Our PD method clearly outputs better detection results in the case of PGD attack. However, for CW and Boundary attacks, LID has not significantly better results, even though on these two more subtle attack algorithms, both PD and LID are almost not able to differentiate clean vs adversarial inputs.

Overall, these experiments confirm what was already stated for the other experiments in the main paper: our PD method still gives better results than LID, or comparable ones.

\subsection{Using number of points in the Persistence Diagram}
\label{sec:nb_pts_detection}

\begin{figure*}
    \centering
    \subfigure{\includegraphics[width=0.55\linewidth]{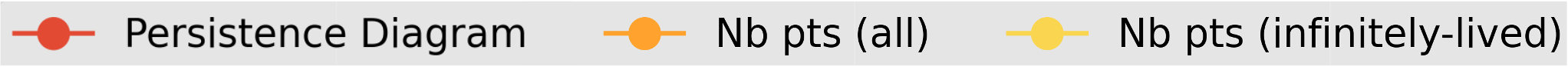}}
    \setcounter{subfigure}{0}
    
    \centering
    \subfigure[LeNet / MNIST]{\includegraphics[width=0.245\textwidth]{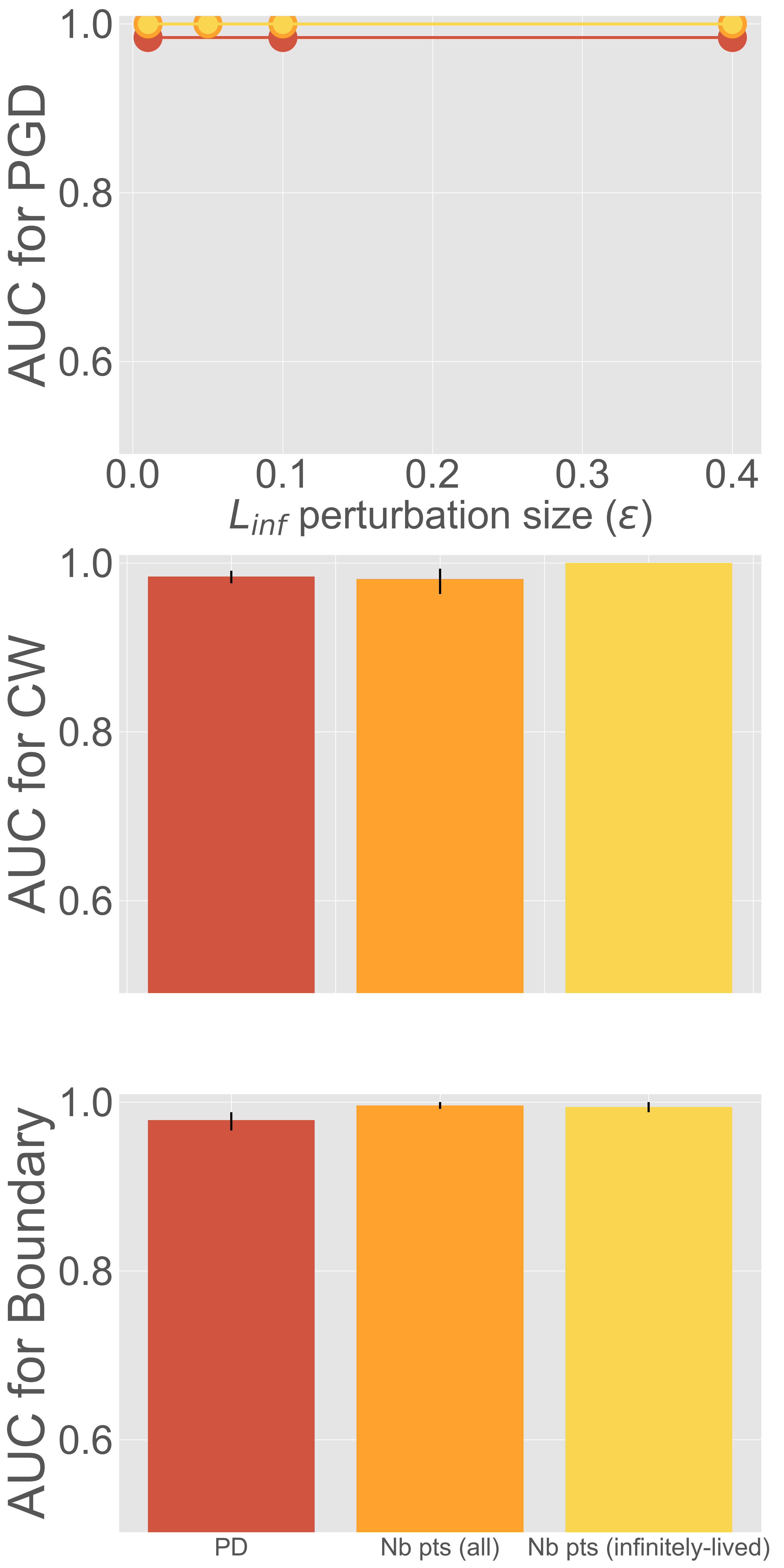}}
    \subfigure[LeNet / Fashion MNIST]{\includegraphics[width=0.245\textwidth]{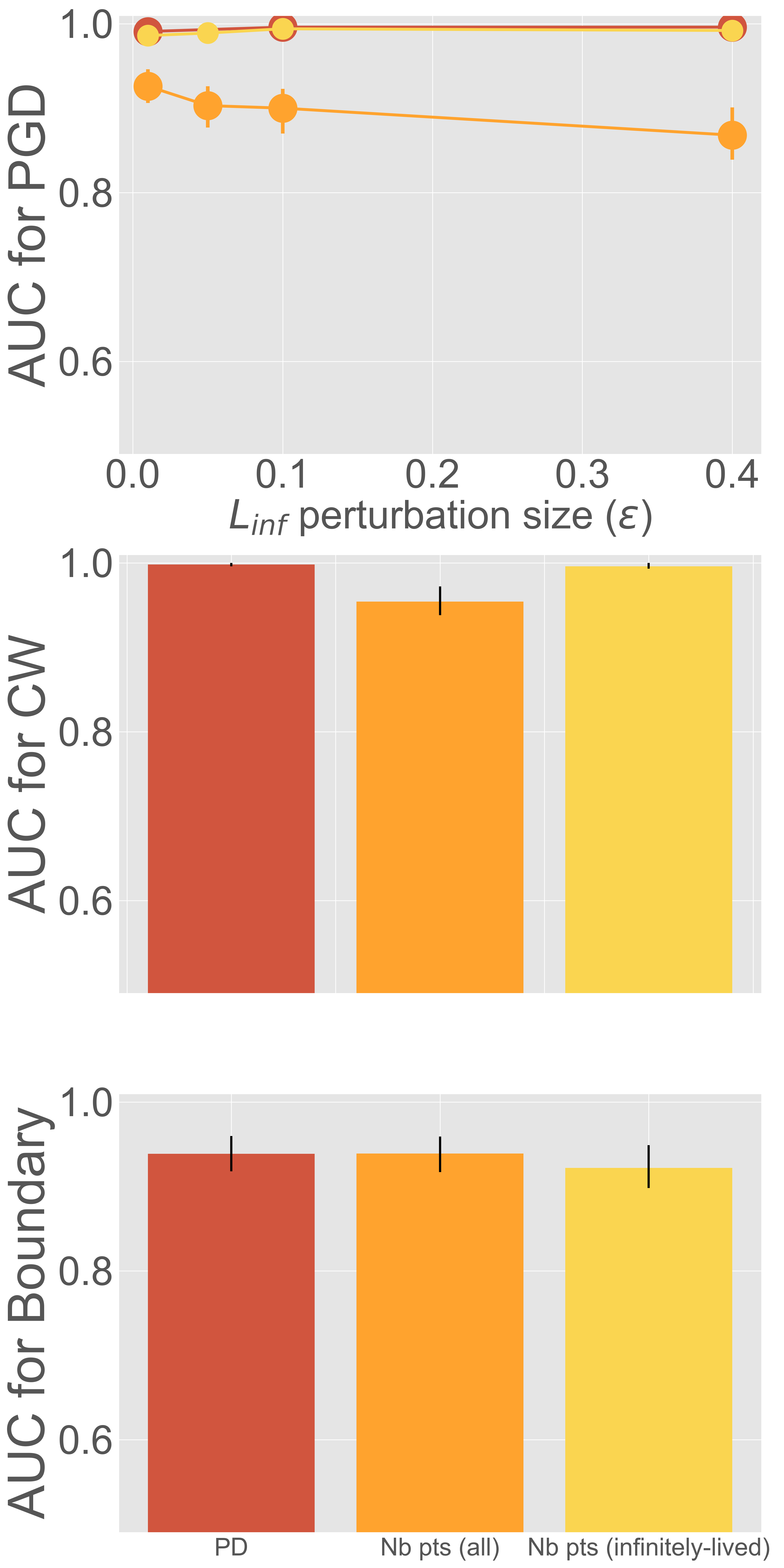}}
    \subfigure[ResNet / CIFAR10]{\includegraphics[width=0.245\textwidth]{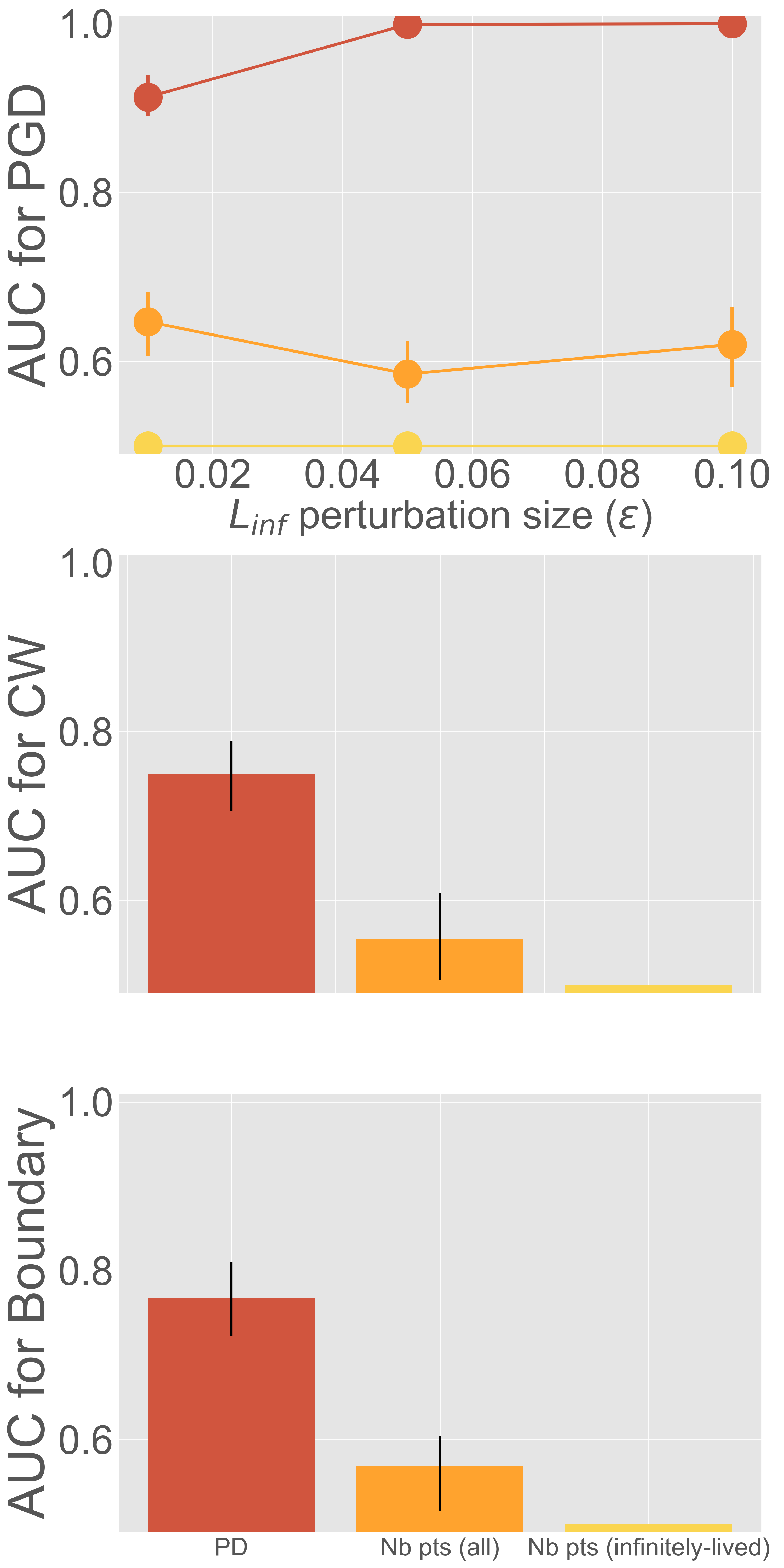}}
\caption{Unsupervised detection results using number of points only.}
\label{fig:nb_pts_dgms_exp}
\end{figure*}

We showed in \cref{sec:preliminary_exp} that using the number of points in PDs can be an efficient strategy to differentiate clean vs adversarial inputs. To emphasize these results, we created two very simple detectors based on the number of points in persistence diagrams (one for all points, one for infinitely-lived points) using an SVM with an RBF kernel. The results are shown in \cref{fig:nb_pts_dgms_exp}. It illustrates the fact that indeed, the number of points in diagrams provides relevant information, even enough to match our PD method in the two simplest settings. When the task is more difficult, however (in CIFAR10 / ResNet setting), it is not enough to yield as good results as when using directly all information from persistence diagrams, like in our PD method.


\section{PD generalizes better than SOTA - Adversarial training experiments.}
\label{sec:adv_training_exp}

\begin{figure*}[h!]
    \centering
    \subfigure[MNIST / LeNet]{\includegraphics[width=0.35\linewidth]{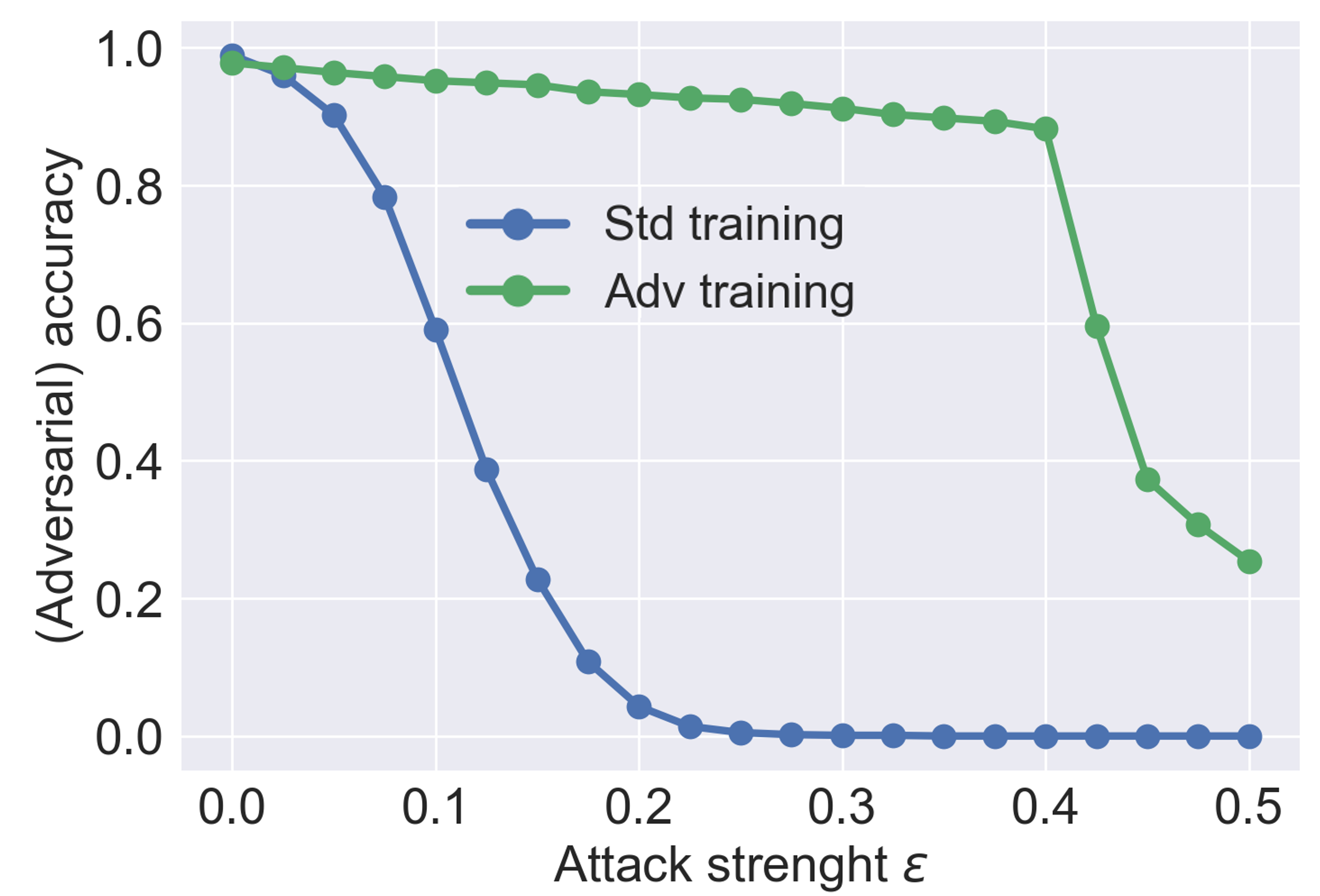}}
    \subfigure[Fashion MNIST / LeNet]{\includegraphics[width=0.35\linewidth]{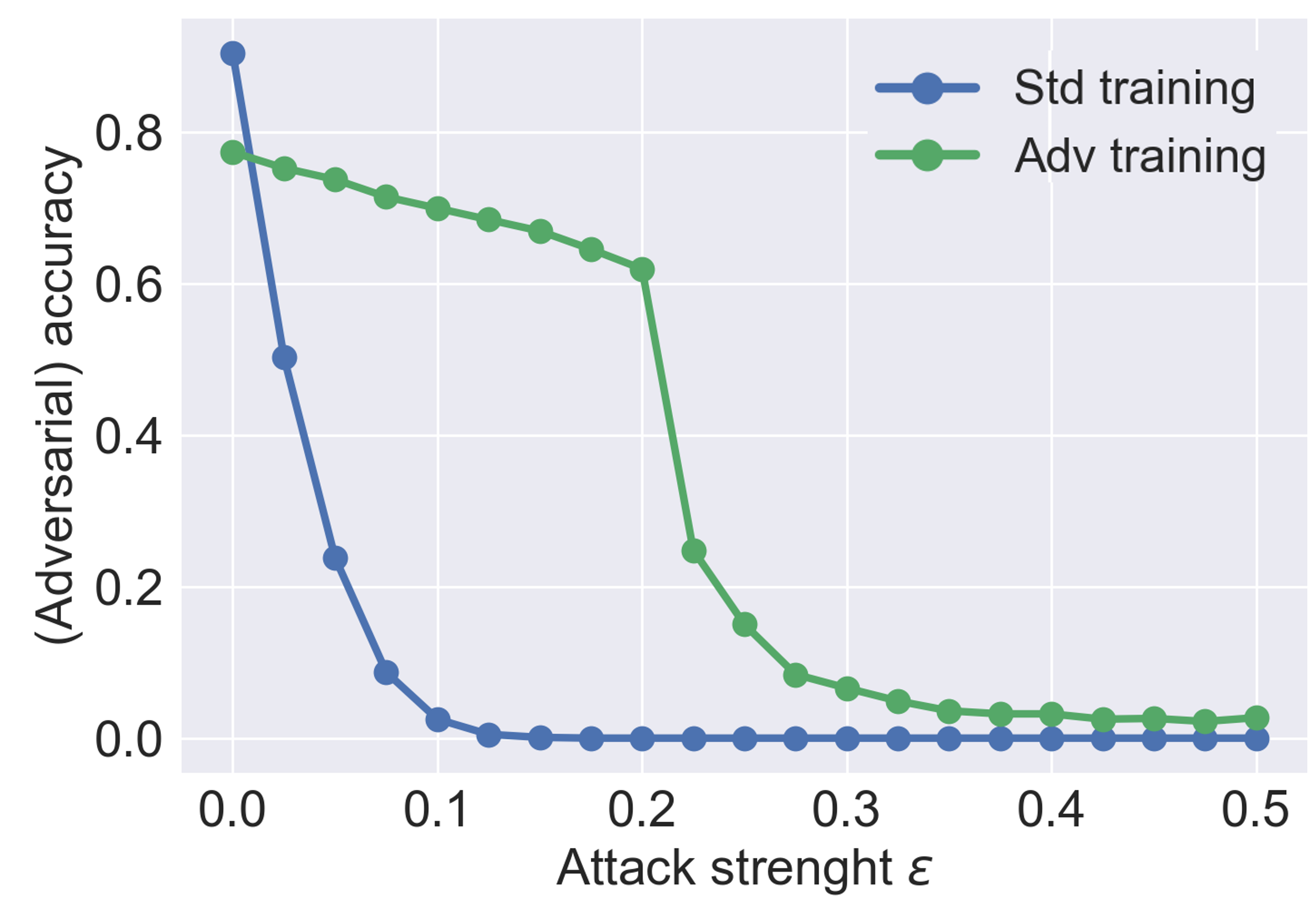}}
\caption{Adversarial accuracy (against PGD) of AT vs standard NNs.}
\label{fig:acc_at_vs_standard}
\end{figure*}

\begin{figure*}
    \centering
    \subfigure{\includegraphics[width=0.5\linewidth]{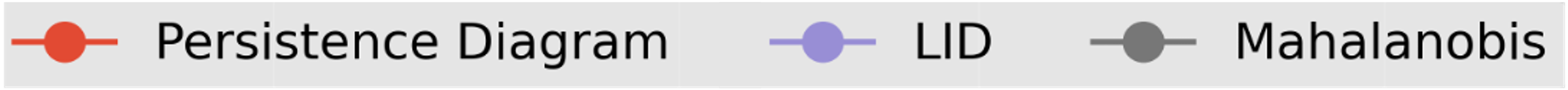}}
    \setcounter{subfigure}{0}
    
    \centering
    \subfigure[LeNet / MNIST]{\includegraphics[width=0.3\textwidth]{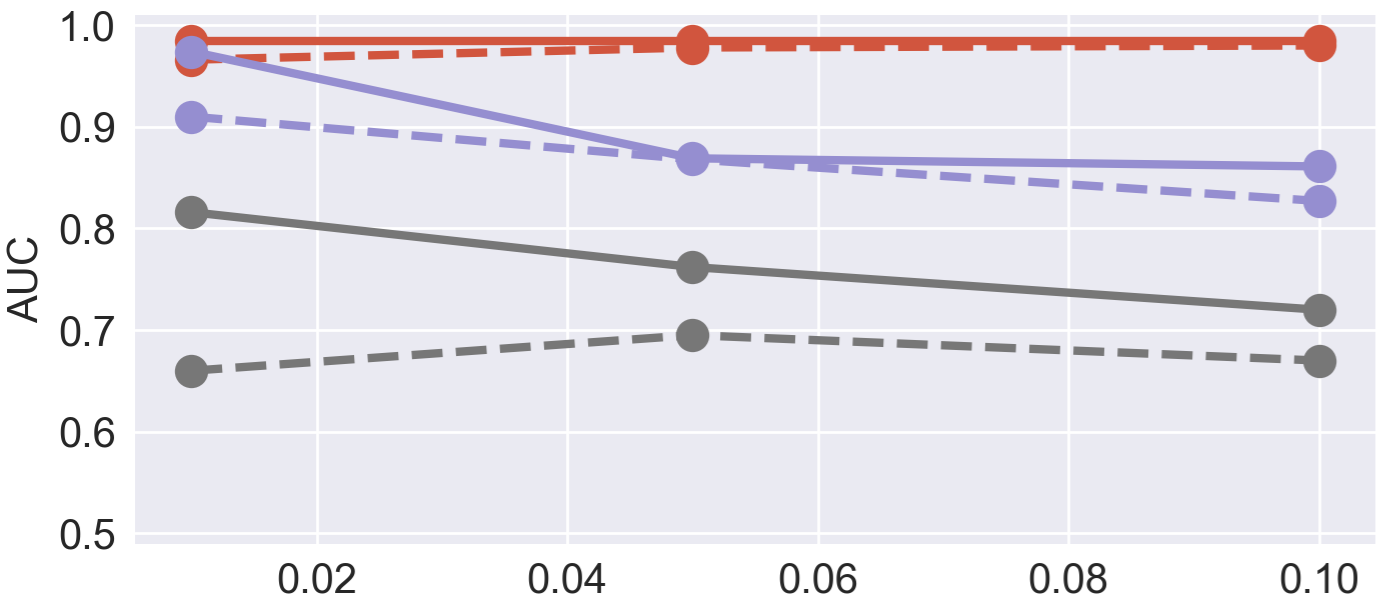}}
    \subfigure[LeNet / Fashion MNIST]{\includegraphics[width=0.3\textwidth]{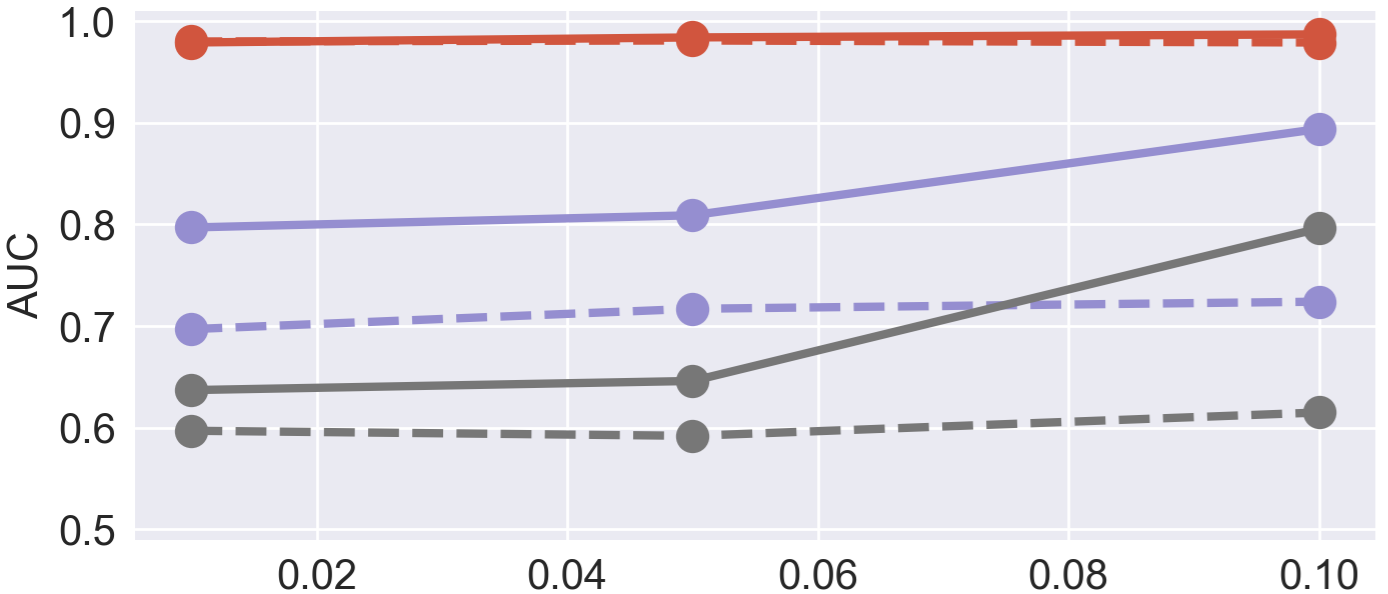}}
\caption{Unsupervised detection results (on PGD) of AT vs standard NNs}
\label{fig:detection_at_vs_std}
\end{figure*}

We illustrated in the main paper the fact that by being a structural method, PD can generalize to all sorts of adversaries. Successful adversaries on adversarially trained (AT) NNs are unusual adversaries by nature, because they can fool a robust model trained to resist the usual adversaries. As such, running our detection methods on AT NNs is a good way to check the generalization ability of said methods: if there is no drop in performance compared to the classical setting, then the method is highly generalizable; if there is one, maybe the method was built on too strong assumptions about adversaries that are not satisfied by all of them.

\cref{fig:acc_at_vs_standard} shows the standard and adversarial accuracy against PGD of the AT NNs compared to the standard ones. \cref{fig:detection_at_vs_std} shows the detection results' discrepancies between standard and AT NNs using PGD attacks, for all methods. Our PD method outputs almost no performance gap, contrary to LID and Mahalanobis, meaning that our method is more general, and that all types of adversaries do leverage under-optimized edges.


\section{Under-optimized edges provide more information than others}
\label{sec:underopt_edges_appendix}

\begin{figure}[h!]
\centering
\includegraphics[width=0.35\textwidth]{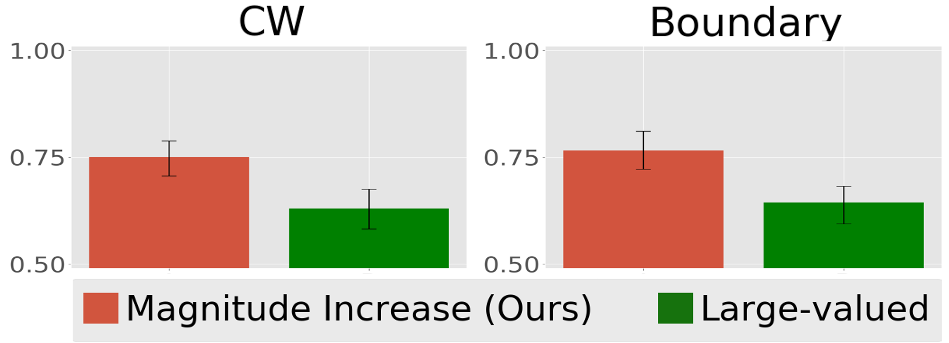}
\caption{Impact of edge-selection methods on AUC (ResNet / CIFAR10).}
\label{fig:edges_selection}
\end{figure}

We provide here an experimental illustration of the impact of edge-selection by comparing the use of under-optimized edges to detect adversarial inputs with our PD method, instead of "well-optimized" edges.
The results shown in \cref{fig:edges_selection} indicate that the detection AUC is better when using under-optimized edges vs well-optimized ones, which also supports our hypothesis stating that these edges contain more information about adversaries.

\section{Pruning and robustness:  and further details}
\label{sec:proof_prop}

\subsection{A theoretical argument: pruning can improve robustness}
\label{subsec:pruning}

We have shown that structural information flow in under-optimized edges are different for clean vs adversarial inputs: these edges represent a vulnerability for NNs. A natural robustification idea would stems from pruning, i.e. exactly removing these under-optimized edges during training. We present a theoretical argument showing how having less active paths, e.g. by pruning, can help robustness.
For an input example $x \in \mathcal X$, let $\mathcal P(x)$ be the set of all weighted paths in the activation graph $G(x,g)$ of $x$ as defined in \cref{sec:induced_graph_threshold}. Each $\alpha \in \mathcal P(x)$ can be identified with a schema
$
u^0(\alpha) \overset{w^1(\alpha)}{\longrightarrow} u^1(\alpha) \overset{w^2(\alpha)}{\longrightarrow} \ldots \overset{w^{L-1}(\alpha)}{\longrightarrow} u^L(\alpha),
$
where $u^l(\alpha) \in [\![1,n_l]\!]$ is the index of the neuron through which the path traverses the $l$th layer of the network, and $w^l(\alpha)$ is the weight of edge weight connecting the former neuron to the next neuron on the path. The subset $\mathcal A(x)$ of paths which are active for the input example $x$ is given by
$\mathcal A(x) := \{\alpha \in \mathcal P(x) \mid w^l(\alpha) \ne 0\;\forall l \in [\![1,L]\!]\}$. Information from input to output only flows along such paths. Finally, let 
$W(\alpha) := \Pi_{l=1}^L(W_l)_{u^{l-1}(\alpha),u^l(\alpha)}$
be the product of all the parameters of the NN along the path $\alpha$. 
We have the following result (the proof is in \cref{sec:proof_prop}).

\begin{restatable}{proposition}{lemma}
\label{sec:lemma}
For every class label $k \in [\![1,K]\!]$ and input feature index $j \in [\![1,n_0]\!]$, we have:
$\frac{\partial [g(x)]_k}{\partial x_j} = \sum_{\alpha} W(\alpha),$ where the sum runs over all active paths $\alpha \in \mathcal A(x)$ such that $u^0(\alpha)=j$ and $u^L(\alpha)=k$, i.e., active paths which start at the $j$th input neuron and end at the $k$th output neuron.
\label{lm:pathsum}
\end{restatable}

Note that the (Frobenius) norm of the jacobian matrix $J(x) = (\frac{\partial g(x)_k}{\partial x_j})_{j,k}$ is a proxy for the robustness to perturbations on input $x$, as it is related to the distance to the closest adversarial example for $x$ (see~\cite{jakubovitz2018improving} and \cref{sec:proof_prop}).
Thus, decreasing this sum improves robustness: we could 1) decrease/remove large $W(\alpha)$ (but it would likely hinder the standard accuracy) or 2) reduce the cardinality of $\mathcal A(x)$, i.e., have very few active paths: this can be achieved by pruning a NN and suggests that under-optimized edges may be a problem for robustness because of their quantity.

\paragraph{Illustration.}
\label{sec:illustration_pruning}

\begin{wrapfigure}[10]{l}{0.47\textwidth}
\vspace{-0.4cm}
    \centering
    \includegraphics[width=0.46\textwidth]{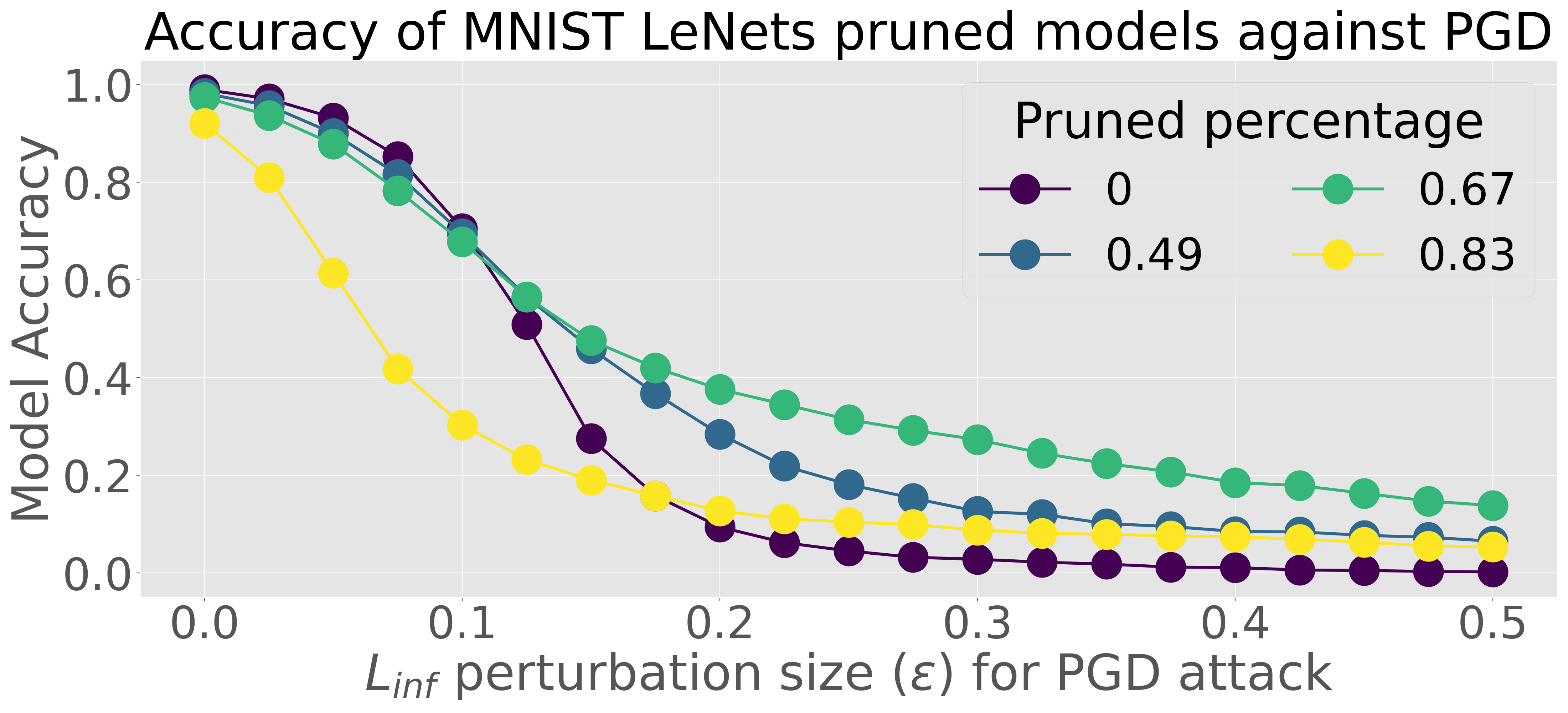}
\caption{Adversarial accuracy of pruned MNIST LeNet models against PGD.}
\label{fig:pruned_adv_acc}
\end{wrapfigure}Some works have focused on the link between adversarial robustness and sparsity~\citep{guo2018sparse, wang2018adversarial, wang2020achieving} but the conclusion remains unclear. We pruned an MNIST LeNet model (following \cite{frankle2018lottery}'s protocol and our definition of under-optimized edges and ran PGD attacks to measure each model's adversarial accuracy. \cref{fig:pruned_adv_acc} shows that some degree of under-optimized edges pruning might be helpful for adversarial robustness (e.g. 67\% seems to be desirable).

\subsection{Proof of \cref{sec:lemma}}

Let us first recall the \cref{sec:lemma}: \lemma*

Note that it holds for the ReLU activation.

\begin{proof}
Let $z_l := g_l(x) \in \mathbb R^{n_l}$ be the output of the $l$th layer of the neural network. Note that $z_{l} = \sigma_l(W_lz_{l-1})$.
By the chain rule, we have
\begin{eqnarray}
\frac{\partial [g(x)]_k}{\partial x_j} = \sum_{k'=1}^{n_{L-1}}\frac{\partial [z_L]_k}{\partial [z_{L-1}]_{k'}}\cdot \frac{\partial [z_{L-1}]_{k'}}{\partial x_j}.
\label{eq:chainrule}
\end{eqnarray}
On the other hand, for ReLU activation we have (still via the chain rule)
$$
\frac{\partial [z_L]_k}{\partial [z_{L-1}]_{k'}} = [W_l]_{k,k'}\sigma'(W_lz_{l-1})=
[W_l]_{k,k'}\begin{cases}1,&\mbox{if }[W_l]_k^\top z_{k'} > 0,\\0,&\mbox{ else.}\end{cases}
$$
Thus the claim follows directly from \eqref{eq:chainrule} by recurring on the depth $L$.
\end{proof}

\subsection{About the Jacobian matrix and its relation with robustness.} In \cite{sokolic2017robust}, authors have shown that the Frobenius norm of the Jacobian matrix is related to the generalization error: regularizing it induces smaller generalization errors. Following this work, \cite{jakubovitz2018improving} have linked the Jacobian matrix to adversarial robustness. For an input $x$, the Froebenius norm of the Jacobian matrix at point $x$ is related to the distance to its closest adversarial example (more precisely, their proposition 3 shows it is an upper bound for the $L_2$-norm of distance to the closest adversary of $x$): minimizing this norm thus leads to improved robustness.